\documentclass{article}

\usepackage{PRIMEarxiv}

\usepackage[utf8]{inputenc}
\usepackage[T1]{fontenc}
\usepackage[hidelinks]{hyperref}
\usepackage{url}
\usepackage{booktabs}
\usepackage{amsfonts}
\usepackage{nicefrac}
\usepackage{microtype}
\usepackage{fancyhdr}
\usepackage{graphicx}
\usepackage{xcolor}
\usepackage{amsmath}
\usepackage{amssymb}
\usepackage{pifont}
\usepackage{multirow}
\usepackage{listings}
\usepackage{subcaption}
\usepackage{tikz}
\usetikzlibrary{arrows.meta,positioning,fit,calc}

\lstdefinestyle{prompt}{
    backgroundcolor=\color{gray!10},
    basicstyle=\ttfamily,
    frame=single,
    breaklines=true,
    captionpos=b,
    breakindent=0pt,
}

\newcommand{\cmark}{\ding{51}}%
\newcommand{\xmark}{\textcolor{red}{\ding{55}}}%
\newcommand{\hflogo}{\raisebox{-0.2ex}{\includegraphics[height=1.2em]{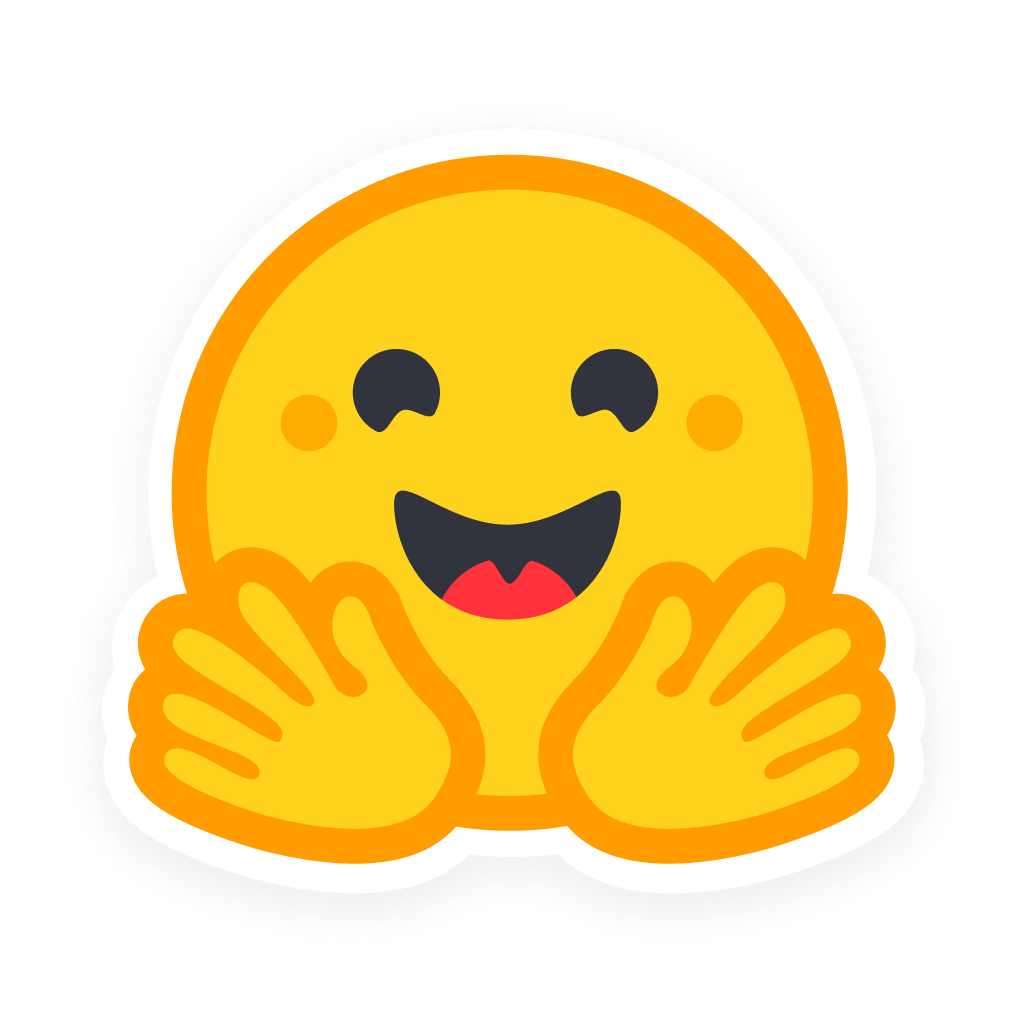}}}
\newcommand{\hfmodel}[1]{%
  \href{https://huggingface.co/rootsautomation/#1}{\texttt{#1}}%
}
\newcommand{\modelline}{%
  \hflogo\ \textbf{Models \& Demo:}~\hfmodel{GutenOCR-3B} \textbar\ \hfmodel{GutenOCR-7B} \textbar\ 
  \href{https://ocr.roots.ai/}{\texttt{Demo}} \textbar\
  \textbf{Code:}~\href{https://github.com/Roots-Automation/GutenOCR}{\texttt{GitHub}}%
}

\graphicspath{{media/}}

\newcommand{\norm}[1]{\operatorname{norm}\!\left(#1\right)}
\newcommand{\lev}{d_{\text{lev}}}
\newcommand{\len}[1]{\left\lvert #1 \right\rvert}
\newcommand{\words}[1]{\left\lvert #1 \right\rvert_{\text{words}}}
\newcommand{\IoU}{\operatorname{IoU}}
\newcommand{\area}{\operatorname{area}}

\newcommand{\tp}{\mathrm{TP}}
\newcommand{\fp}{\mathrm{FP}}
\newcommand{\fn}{\mathrm{FN}}

\title{%
  \begin{minipage}{0.15\textwidth}
    \centering
    \includegraphics[width=1.8cm,height=1.8cm]{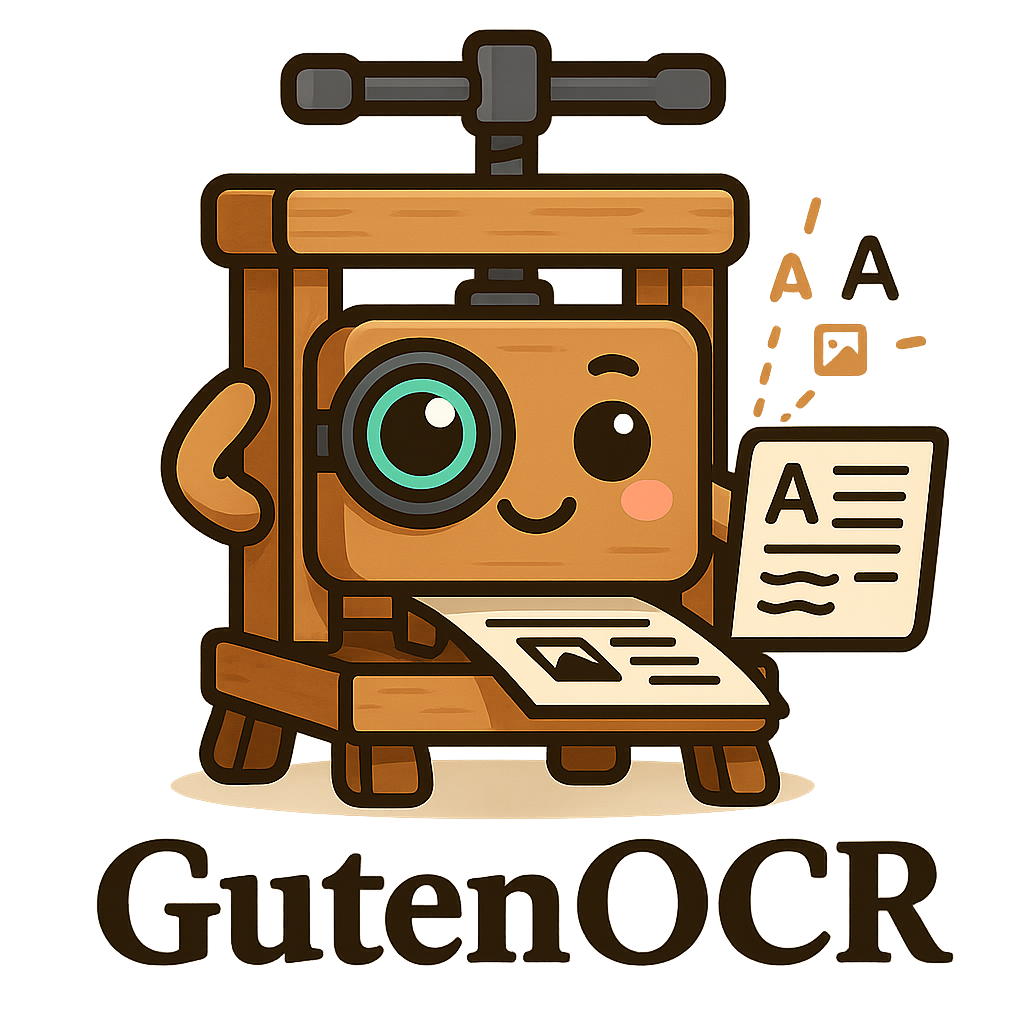}
  \end{minipage}\hfill
  \begin{minipage}{0.85\textwidth}
    \raggedright
    GutenOCR: A Grounded Vision--Language Front-End for Documents
  \end{minipage}
}

\author{
  Hunter Heidenreich, Ben Elliott, Olivia Dinica, Yosheb Getachew \\
  Roots.ai \\
  \texttt{ai-ml@roots.ai} \\
}

\begin{document}
\maketitle
\vspace{-3.75em}
\begin{center}
{\small \modelline}
\end{center}
\vspace{1.75em}

\begin{abstract}
GutenOCR is a family of grounded OCR front-ends obtained 
by fine-tuning Qwen2.5-VL-3B and Qwen2.5-VL-7B. 
The resulting single-checkpoint vision--language models expose 
reading, detection, and grounding through a unified, prompt-based interface.
Trained on business documents, scientific articles, 
and synthetic grounding data, the models support full-page and localized reading
with line- and paragraph-level bounding boxes 
and conditional ``where is x?'' queries. 
We introduce a grounded OCR evaluation protocol and show that GutenOCR-7B 
more than doubles the composite grounded OCR score of its Qwen2.5-VL-7B backbone 
on 10.5K held-out business and scientific pages (0.40$\rightarrow$0.82).
On Fox and OmniDocBench v1.5, our approach substantially improves region- and line-level OCR as well as text-detection recall, but reveals trade-offs in page-level linearization, color-guided OCR, and formula-heavy layouts.
\end{abstract}

\begin{figure}[htb]
    \centering
    \begin{minipage}[c]{0.7\linewidth}
        \includegraphics[width=\linewidth]{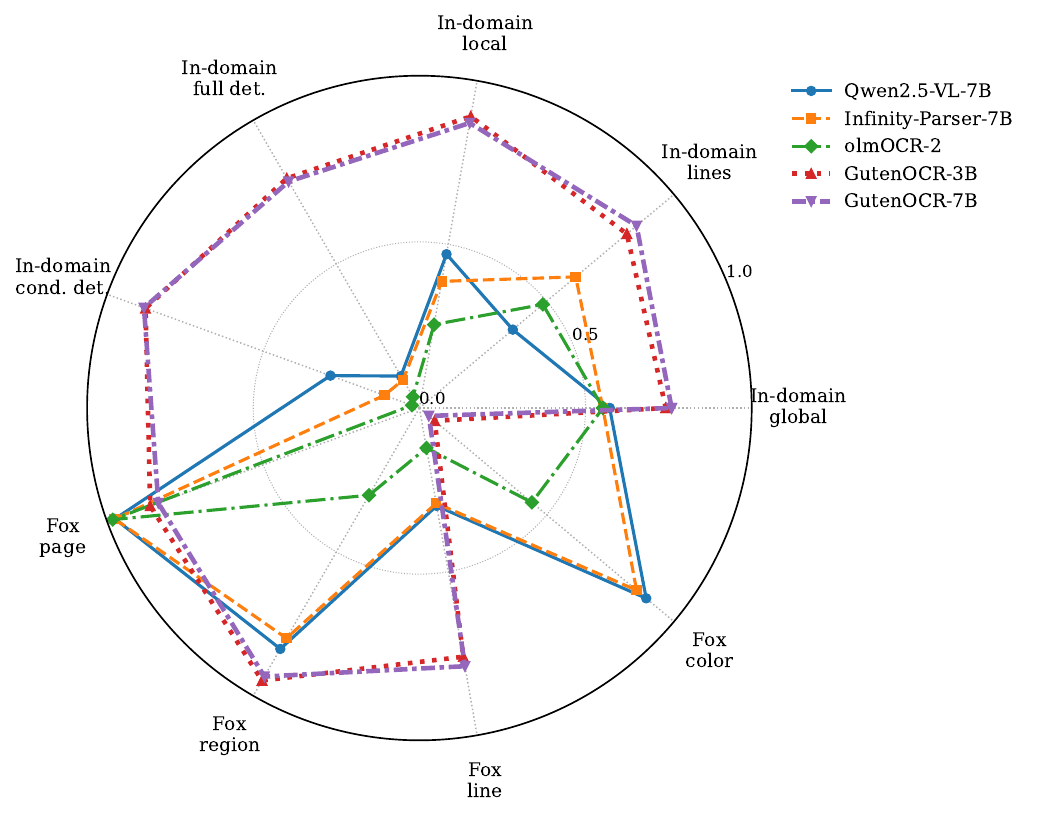}
    \end{minipage}\hfill
    \begin{minipage}[c]{0.28\linewidth}
        \captionof{figure}{
            Capability profile of GutenOCR-3B/7B, their Qwen2.5-VL backbones, and Qwen2.5-based OCR baselines on nine grounded OCR tasks spanning our in-domain suite and Fox. Each spoke shows a raw score in $[0,1]$ (higher is better): in-domain page-, line-, and local-reading tasks are scored as $1-\mathrm{CER}$, full and conditional detection as F1, and Fox page-, region-, line-, and color-guided OCR as $1-\mathrm{CER}$. GutenOCR substantially improves detection, fine-grained reading, and Fox region/line OCR, while trading off some page-level and color-guided performance relative to OCR-specialized baselines.
        }
        \label{fig:hero}
    \end{minipage}
\end{figure}

\tableofcontents

\section{Introduction}
\label{sec:intro}

Document images remain a primary medium for business transactions, 
scientific communication, and automated workflows. 
Optical character recognition (OCR) is therefore the front door 
for much of the text consumed by large language models (LLMs) 
and retrieval-augmented generation (RAG) systems: 
any content that OCR fails to capture is effectively invisible 
to downstream, text-only systems.

Two main approaches currently shape document OCR. 
Classical pipelines---with distinct stages for detection, recognition, and 
layout analysis---offer explicit text grounding and a controllable reading order, 
but are brittle on complex layouts and difficult to adapt to new domains or tasks~\cite{TessOverview,cui2025paddleocr30technicalreport,amazon_textractor,google_cloud_ocr}.
``OCR-free'' vision--language models (VLMs), by contrast, now dominate many 
document-understanding benchmarks~\cite{kim2022donut,wei2024generalocrtheoryocr20,poznanski2025olmocr2unittest,wang2025infinityparserlayoutaware},
scaling to large datasets, handling diverse visual signals, 
and operating in complex domains (e.g., chemistry, music, mathematics).
However, they typically treat text as a latent variable: page-level transcripts are implicit,
token-level grounding is at best partial, and reading behavior is controllable 
only indirectly via prompts.
We refer to the OCR component that downstream systems actually interact with as a 
\emph{grounded OCR front-end}: a model that (i) produces page transcripts, 
(ii) attaches each token or span to 2D bounding boxes, and 
(iii) can be prompted to read arbitrary regions on the page.

This lack of an explicit grounded front-end is particularly acute in production settings, 
where downstream systems need both high-quality text and fine-grained control 
over \emph{how} it is read. 
They require precise local reading, explicit links between tokens and pixels, 
and representations that can be tailored (e.g., whitespace-sensitive, layout-aware ordering) 
for extraction, filtering, and RAG. 
Non-grounded page-to-Markdown outputs make human oversight costly: 
reviewers must read both the full page and the generated transcript end-to-end, 
which is slow and error-prone. 
In contrast, grounded outputs make errors easier to spot and correct: 
hallucinated content appears as boxes over empty regions, 
missing text manifests as gaps in box coverage, 
and misrecognized spans are localized to specific boxes. 
Grounding therefore directly supports human-in-the-loop verification 
and active-learning workflows.

Existing OCR-specialized VLMs and end-to-end parsers~\cite{poznanski2025olmocr2unittest,wang2025infinityparserlayoutaware,wei2025deepseekocrcontextsopticalcompression,cui2025paddleocrvlboostingmultilingualdocument}
deliver strong page-to-Markdown quality, 
but are not designed as grounded front-ends: they often couple OCR with task-specific decoding formats, 
expose limited or unstable token--box alignment, 
and offer only coarse control over where and how the page is read. 
When grounded modes are provided, they are frequently tied to bespoke prompt conventions 
or rigid output schemas that downstream systems must accommodate. 
Classical OCR engines, in contrast, expose modular, stable APIs for detection, recognition, 
and layout analysis that can be flexibly composed inside larger business workflows. 
Rather than forcing legacy systems to retrofit around model-specific formats, 
we seek VLM-based OCR systems that offer equally versatile, API-like interfaces.

We revisit OCR as a grounded front-end problem and introduce \textbf{GutenOCR}, 
a VLM-based OCR system trained to behave like a traditional OCR pipeline 
while retaining the flexibility of contemporary vision--language architectures~\cite{bai2025qwen25vltechnicalreport,bai2025qwen3vltechnicalreport}. 
A single checkpoint supports full-page reading and detection, localized reading, and conditional detection, 
and can return either plain transcripts or structured line- and paragraph-level outputs with bounding boxes. 
By releasing the model and training recipe, we aim to establish an open baseline for VLM-based OCR 
that combines the strengths of classical pipelines with the expressivity and scalability 
of general-purpose multimodal models, providing a drop-in, human-verifiable front-end 
for downstream extraction and RAG systems.

In summary, this work makes five contributions:
\begin{itemize}
    \item \textbf{Unified grounded OCR front-end.} We present GutenOCR, a family of 3B and 7B VLM checkpoints that,
    via prompt-specified input--output schemas, support full-page reading and detection,
    conditional detection, and localized reading. The models return either linear transcripts or structured
    line- and paragraph-level outputs with bounding boxes, integrating cleanly with existing business workflows
    and human-in-the-loop verification pipelines.

    \item \textbf{OCR specialization recipe for general-purpose VLMs.} 
    We provide a training recipe on a public data mixture that turn Qwen2.5-VL-3B/7B~\cite{bai2025qwen25vltechnicalreport} 
    into grounded OCR-specialized models without modifying the tokenizer, adding adapters, or freezing modules, 
    while supporting long-context page transcripts (up to 8k--16k tokens).

    \item \textbf{Metrics and protocol for grounded OCR.} We define a metric suite and evaluation protocol 
    that jointly measure text accuracy, detection quality, and page-level fidelity across task families 
    using CER, WER, F1@0.5, mCER@0.5, and $\text{CER}_\text{e2e}$. 
    These metrics separate pure recognition errors from localization and reading-order failures, 
    avoiding ``good reader, bad paginator'' pathologies.

    \item \textbf{Empirical analysis across in-domain and external benchmarks.} 
    On 10.5K held-out business and scientific pages, GutenOCR-7B substantially improves a composite grounded OCR score, 
    localized-reading CER, and line-detection F1 over its Qwen2.5-VL-7B backbone. 
    On Fox~\cite{liu2024fox} and OmniDocBench v1.5~\cite{ouyang2025omnidocbenchbenchmarkingdiversepdf}, 
    GutenOCR significantly reduces region- and line-level OCR CER and boosts text-detection recall, 
    while revealing clear failure modes in formula-heavy and color-guided OCR settings (see Figure~\ref{fig:hero}).
    \item \textbf{Reproducible open release.} We release our training and evaluation code, provide full
    data provenance (using only publicly available sources), and publish open model weights under a license compatible with
the upstream data and model licenses. Our training code can be found at \url{https://github.com/Roots-Automation/GutenOCR}.
\end{itemize}

\section{GutenOCR Overview}

\begin{figure}[htb]
    \centering
    \begin{tikzpicture}[
        >=Stealth,
        panel/.style={inner sep=0pt},
        box/.style={draw, rounded corners=2pt, minimum width=2.6cm,
                    minimum height=0.9cm, align=center},
        axislabel/.style={font=\small},
        arrowlabel/.style={font=\small, inner sep=1pt, fill=white, rounded corners=1pt}
    ]
    
    
    \node[panel] (page) at (0,-2)
        {\includegraphics[width=5.0cm]{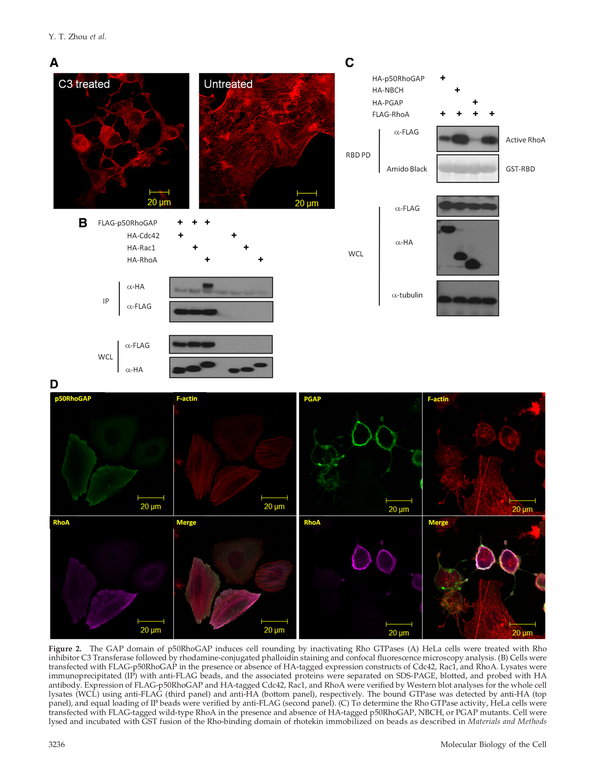}};
    
    \node[panel] (model) at (0,4.0)
        {\includegraphics[width=2.6cm]{media/guten-mascot.png}};
    
    \node[panel] (full-det)  at (-6, 4.0)
        {\includegraphics[width=5.0cm]{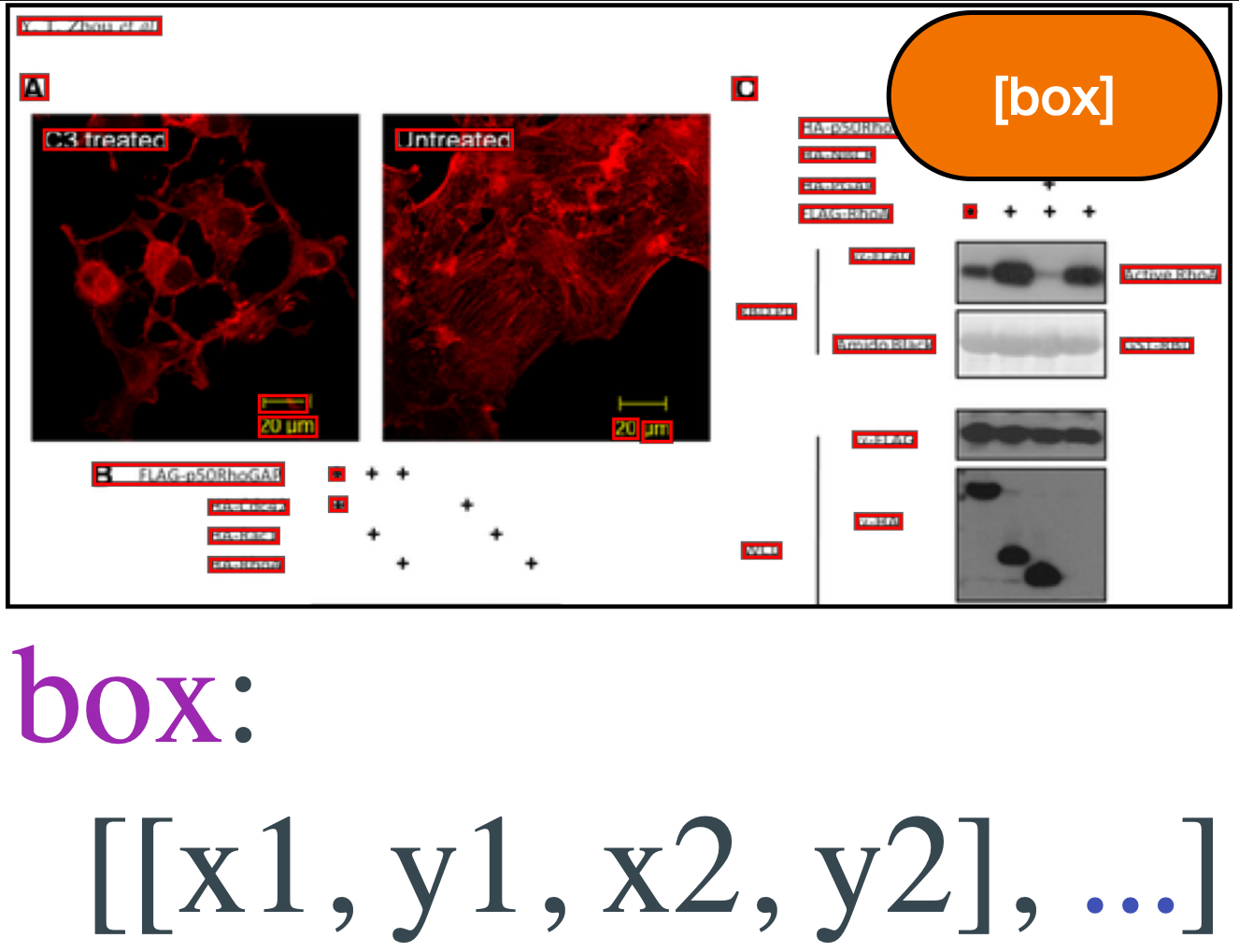}};
    \node[panel] (full-read) at ( 6, 4.0)
        {\includegraphics[width=5.0cm]{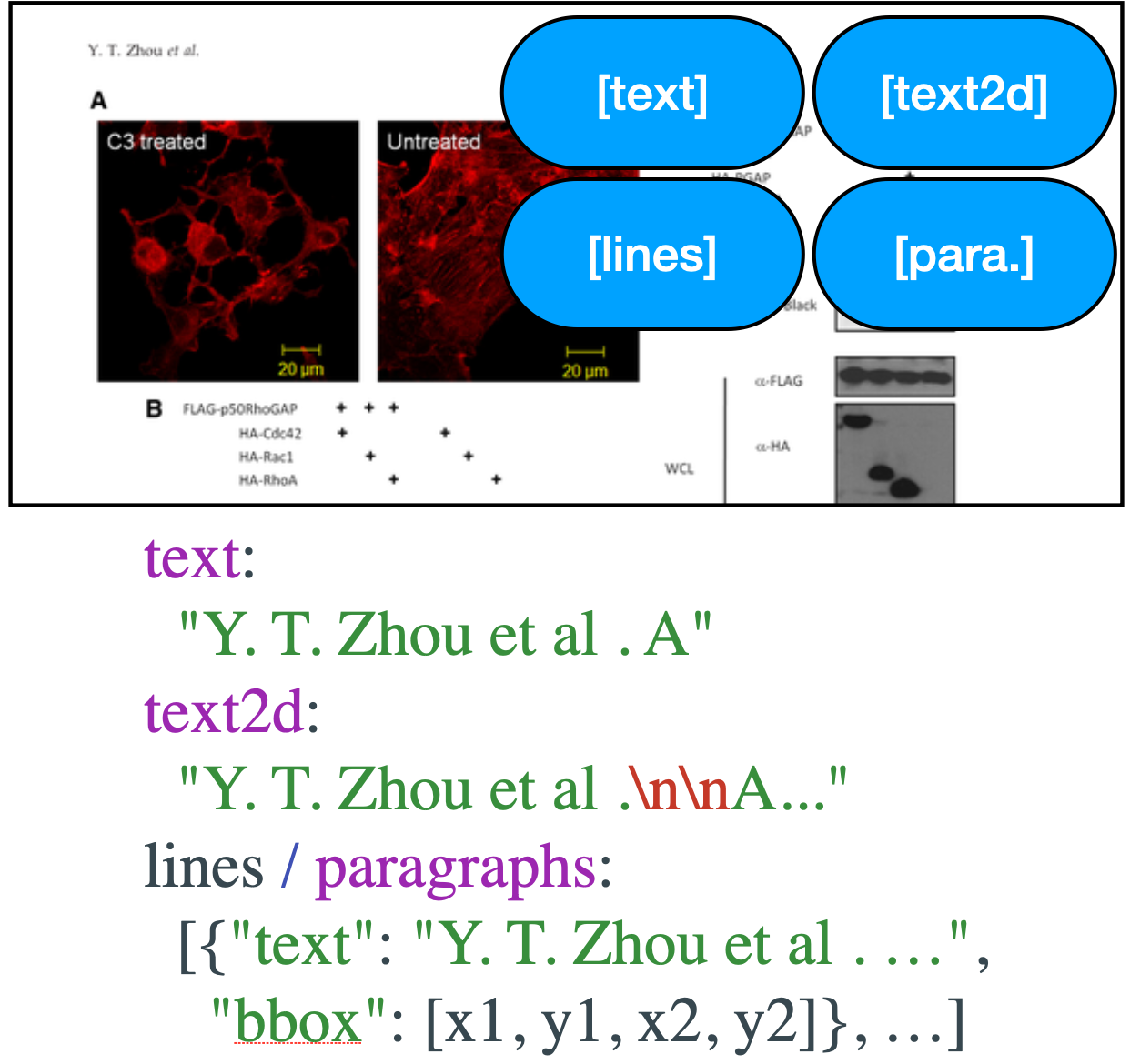}};
    \node[panel] (tgt-det)   at (-6,-3.0)
        {\includegraphics[width=5.0cm]{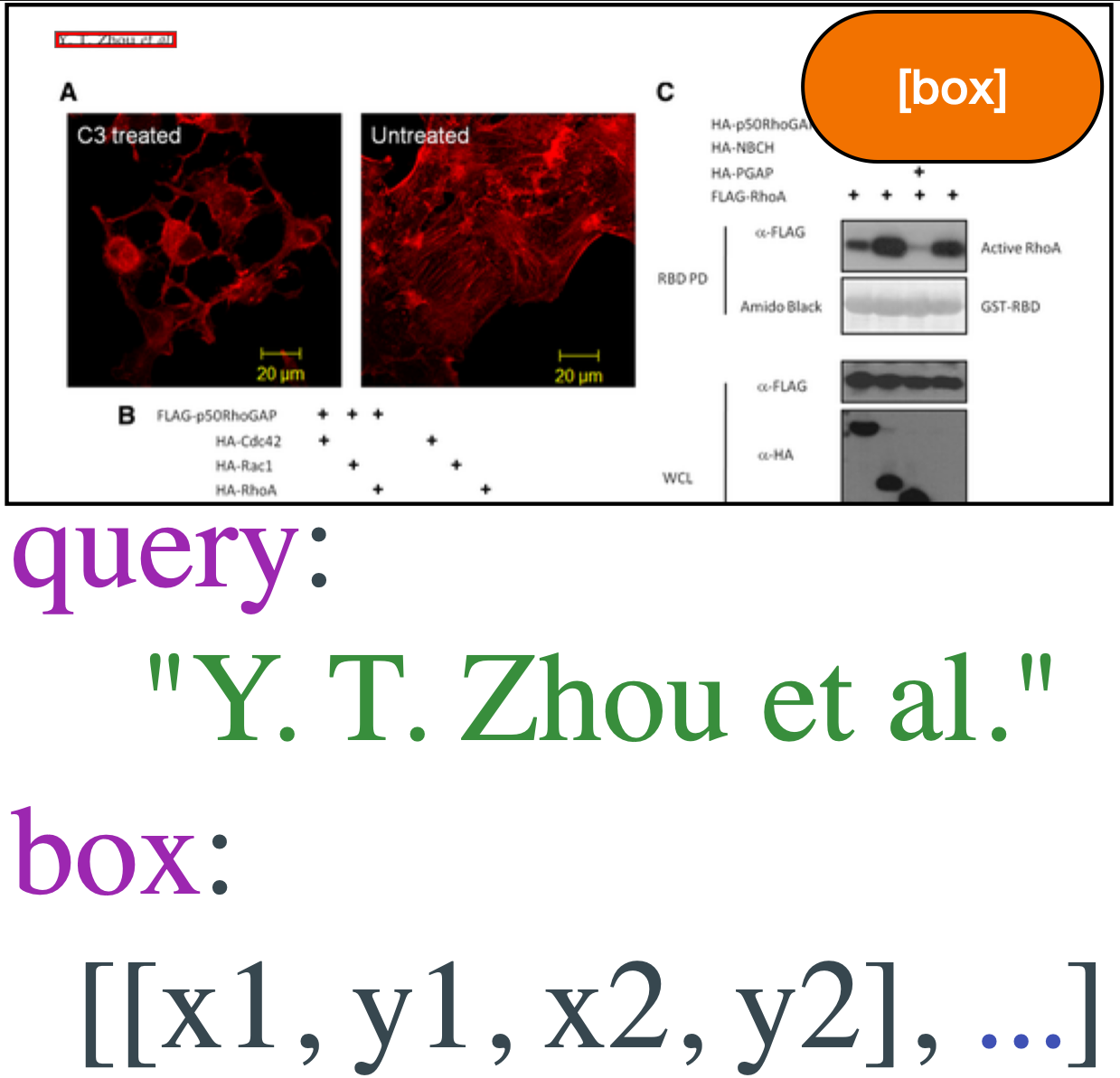}};
    \node[panel] (tgt-read)  at ( 6,-3.0)
        {\includegraphics[width=5.0cm]{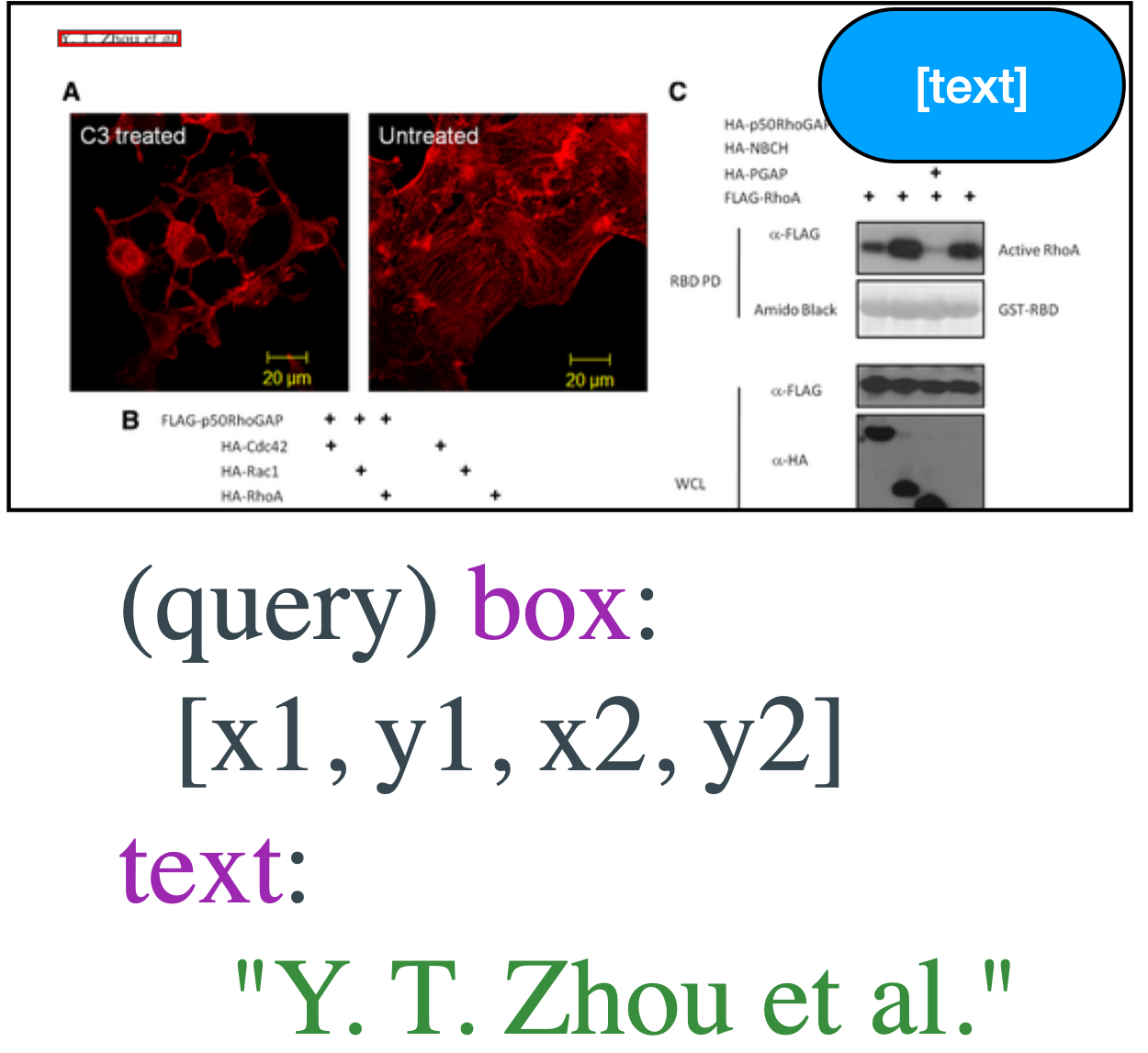}};

    
    \draw[->] (page.north) -- (model.south);
    
    \draw[->] (model.west) -- node[arrowlabel, above]
        {Detect all text} (full-det.east);
    \draw[->] (model.east) -- node[arrowlabel, above]
        {Read all text} (full-read.west);
    
    \draw[->] (model.south west) -- node[arrowlabel, above, sloped]
        {Image + query $q$ $\to$ BOX$(q)$} (tgt-det.north east);
    
    \draw[->] (model.south east) -- node[arrowlabel, above, sloped]
        {Image + box $b$ $\to$ text} (tgt-read.north west);
    
    \end{tikzpicture}
    
    \caption{
        Unified GutenOCR interface. A single vision--language model consumes a page image,
        optionally conditioned on a text query $q$ or bounding box $b$, and serves multiple OCR-style
        tasks (reading, detection, conditional detection, localized reading) through prompt-specified
        input and output schemas.
        Sub-panels depict each task family: 
        full page detection (top-left), 
        full page reading (top-right),
        local reading (bottom-right),
        and conditional detection (bottom-left).
        Sample page from \cite{Zhou2010BCH}.
    }
    \label{fig:rootsocr-interface}
\end{figure}

\subsection{Model Overview}

GutenOCR is a family of grounded OCR front-ends as defined in Section~\ref{sec:intro}. Each model is a single-checkpoint vision--language system, obtained by fine-tuning Qwen2.5-VL~\cite{bai2025qwen25vltechnicalreport}, that exposes classical OCR stages and pointer-like behaviors (for example, ``Where is $x$ in this document?'') via prompts.
Conceptually, GutenOCR acts as a thin, API-like layer between page images and downstream systems, exposing a small set of stable OCR primitives (reading, detection, grounding) that can be composed into diverse document workflows.

As depicted in Figure~\ref{fig:rootsocr-interface},
each GutenOCR model takes a single document page image as input, optionally
augmented with a text query (for conditional detection) or a bounding box
(for localized reading), and produces either
(i) plain transcripts (\texttt{text}, \texttt{text2d}) or
(ii) structured outputs with line- or paragraph-level bounding boxes
(\texttt{lines}, \texttt{paragraphs}), depending on the requested mode.
All task variants share the same backbone, tokenizer, and prompting interface:
the ``pipeline'' stages of a traditional OCR system (reading, detection,
grounding) are exposed as different input--output schemas of a single model,
rather than as separate components.

\subsection{Task Families and Interfaces}
\label{sec:tasks}

\begin{table}[htb]
\centering\small
\begin{tabular}{c c p{2.5cm} p{6cm}}
\toprule
\textbf{Task} & \textbf{Input} & \textbf{Output} & \textbf{Example prompt} \\
\midrule
 & & \texttt{text} &
Read all text in the image and return a single plain-text transcription.\\
\cmidrule{4-4}
\multirow{3}{*}{Reading} &  \multirow{3}{*}{Image}& \texttt{text2d} &
Return a layout-sensitive \texttt{text2d} representation of the image.\\
\cmidrule{4-4}
 & & \texttt{lines} &
Extract line-level JSON objects with \texttt{text} and \([x_1,y_1,x_2,y_2]\).\\
\cmidrule{4-4}
 & & \texttt{paragraphs} &
Extract paragraph-level JSON objects with \texttt{text} and \([x_1,y_1,x_2,y_2]\).\\
\midrule
 &  & &
Detect all LINES in the image and return a JSON array of \([x_1,y_1,x_2,y_2]\) boxes.\\
\cmidrule{4-4}
\multirow{2}{*}{Detection}  & \multirow{2}{*}{Image} & \multirow{2}{*}{\texttt{BOX}} &
Detect all PARAGRAPHS in the image and return a JSON array of boxes.\\
\cmidrule{4-4}
 &  & &
Detect all math regions in the image and return their boxes as JSON.\\
\midrule
Conditional detection & Image + text & \texttt{BOX} &
Given a query string ``$x$'', return boxes for all matching lines in the image.\\
\midrule
Localized reading & Image + box & \texttt{lines\_text}, \texttt{paragraphs\_text} &
Given a bounding box \([x_1,y_1,x_2,y_2]\), read only the text inside that region.\\
\bottomrule
\end{tabular}
\caption{Canonical task families, inputs, outputs, and prompts.}
\label{tab:tasks}
\end{table}

We expose four task families through a unified, prompt-based interface:
full-page reading, full-page detection, conditional detection, and localized
reading.
We use this four-way taxonomy throughout the paper for both training (Section~\ref{sec:training})
and evaluation (Section~\ref{sec:metrics}).
Table~\ref{tab:tasks} gives canonical input--output schemas and example prompts.

For grounded outputs, we use JSON arrays of objects with keys \texttt{text} and \texttt{bbox}, where \texttt{bbox} is $[x_1, y_1, x_2, y_2]$ in pixel coordinates.
For pure detection tasks the output is a JSON array of \texttt{bbox} arrays only.
Full-page reading recognizes text over the entire page and can return either linear transcripts (\texttt{text}) or structured outputs (\texttt{text2d}, \texttt{lines}, \texttt{paragraphs}).
Full-page detection localizes text or other regions of interest without recognition.
Localized reading transcribes user-specified subregions, and conditional detection highlights all instances of a query string.

All outputs of type \texttt{BOX} are axis-aligned bounding boxes in pixel coordinates, represented as \([x_1,y_1,x_2,y_2]\) with \(x_1<x_2\), \(y_1<y_2\) in the original image reference frame.
We do not predict rotated boxes; rotated or skewed text is represented by the axis-aligned box of its minimal enclosing rectangle.
More detailed conventions (e.g., clipping and degenerate boxes) are given in Appendix~\ref{app:interface-geometry}.

\begin{figure}[htb]
  \centering
  \includegraphics[width=0.8\linewidth]{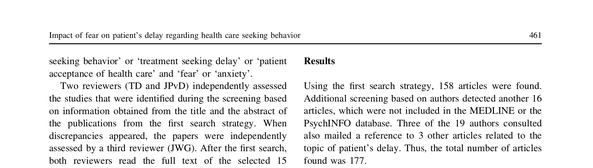}

  \vspace{0.7em}

  \begin{minipage}[t]{\linewidth}
    \centering
    \ttfamily\scriptsize
\begin{verbatim}
Impact of fear on patient's delay regarding health care seeking behavior                                                   461

seeking behavior ' or ' treatment seeking delay ' or ' patient      Results
acceptance of health care ' and ' fear ' or ' anxiety ' .
   Two reviewers ( TD and JPvD ) independently assessed             Using the first search strategy , 158 articles were found .
the studies that were identified during the screening based         Additional screening based on authors detected another 16
on information obtained from the title and the abstract of          articles , which were not included in the MEDLINE or the
\end{verbatim}
  \end{minipage}

  \caption{
    Example of the layout-sensitive \texttt{text2d} representation.
    Long runs of spaces encode horizontal alignment (e.g., right-justified page number),
    while blank lines encode vertical gaps between sections and between the
    main text and the flow diagram.
    Page from \cite{Dubayova2010FearDelay}.
  }
  \label{fig:text2d-example}
\end{figure}

\paragraph{Layout-sensitive \texttt{text2d}.}
The \texttt{text2d} format is a whitespace-preserving, layout-aware
linearization built from line-level boxes and transcripts.
We order lines in a 2D reading order and render them into a notional character
grid, mapping horizontal gaps to spaces and sufficiently large vertical gaps
to blank lines.
The resulting \texttt{text2d} string uses only spaces and newline characters
to encode 2D layout while remaining a plain-string API for downstream systems
(see Figure~\ref{fig:text2d-example}).
The exact construction procedure and its interaction with CER/WER metrics are
described in Appendix~\ref{app:interface-text2d}.

\paragraph{Task-specific semantics.}
Conditional detection tasks take a query string \(q\) and an image as input
and must return bounding boxes for \emph{all} lines whose normalized text
contains \(q\); if the query does not appear, the expected output is an empty
JSON array \texttt{[]}.
Localized reading tasks take a bounding box and an image as input and require the model to transcribe only the text inside that region. We train both line-level and paragraph-level variants, but for localized reading the API always returns a plain-text string (no boxes). We reuse the labels \texttt{lines}/\texttt{paragraphs} only to indicate the expected granularity of the text, not the JSON structure.
Further implementation details are deferred to Appendix~\ref{app:interface-semantics}.

\section{Data \& Training}

\subsection{Data curation}
\label{sec:data}

We construct the training corpus by combining large-scale real-world document
collections with synthetic data targeted at specific grounding behaviors.
Our goal is to train a grounded OCR front-end---rather than a general-purpose
vision--language model---so we bias the mixture toward scanned documents and
synthetic layouts that emphasize line- and paragraph-level geometry, math, and
noisy business forms.
Table~\ref{tab:training-data} summarizes corpus sizes, OCR engines, and
available annotation types; Appendix~\ref{app:data} details preprocessing,
splits, and sampling.

\paragraph{Real sources.}
\emph{OCR-IDL (IDL)}~\cite{biten2022ocridl} covers noisy, heterogeneous
business documents (IDs, invoices, claims) with stamps, handwriting, and
capture artifacts.
\emph{TabMe++ (T++)}~\cite{heidenreich2024pss} offers similar content with a
different OCR backend and layout distribution.
\emph{PubMed-OCR (PMD)}~\cite{heidenreich2025pubmedocr} brings long, multi-column scholarly
articles with equations, Greek/Latin symbols, figure captions, and references.
Training across these sources exposes the model to
(i) distinct page organizations (forms, tables, narrative prose),
(ii) broad typographic and symbol distributions (small fonts, math,
diacritics), and
(iii) varied rasterization noise and reading-order conventions.
IDL and TabMe++ provide line- and word-level transcripts but no paragraph
structure, while PubMed-OCR includes paragraph segmentation in addition to
lines and words.

\paragraph{Synthetic sources.}
We complement real pages with synthetic data that explicitly target grounding
for lines and math.

\begin{itemize}
\item \emph{Grounded \LaTeX{} (GL).}
We mine equation snippets from Wikimedia~\cite{schubotz2025wiki},
normalize them to KaTeX-compatible \LaTeX,
and render them into single-page PDFs at random positions, scales, and
rotations.
Each equation is associated with its tight bounding box in the rendered page,
providing localized math supervision.

\item \emph{SynthDoG grounding (SDG).}
Following prior work~\cite{kim2022donut}, we adapt SynthDoG to
(i) emit line-level bounding boxes,
(ii) sample text from FinePDFs~\cite{kydlicek2025finepdfs}, and
(iii) use word-sensitive wrapping to avoid mid-word line breaks.
These synthetic pages provide dense supervision for line detection and
structured reading.
\end{itemize}

\begin{table}[htb]
    \centering\small
    \begin{tabular}{lrrccccc}
    \toprule
    \textbf{Dataset} & \textbf{Docs} & \textbf{Pages} & \textbf{OCR engine} & \textbf{Paragraphs} & \textbf{Lines} & \textbf{Words} & \textbf{\LaTeX{}} \\
    \midrule
    OCR-IDL             & 4.6 M   & 26.0 M  & Amazon Textract   & \xmark & \cmark & \cmark & \xmark \\
    TabMe++             & 44.8 K  & 122.5 K & Azure OCR         & \xmark & \cmark & \cmark & \xmark \\
    SynthDoG Grounding  & --      & 1.2 M   & Synthetic         & \xmark & \cmark & \xmark & \xmark \\
    Grounded \LaTeX{}   & --      & 3.0 M   & Synthetic         & \xmark & \xmark & \xmark & \cmark \\
    PubMed-OCR          & 209.5 K & 1.5 M   & Google Vision OCR & \cmark & \cmark & \cmark & \xmark \\
    \bottomrule
    \end{tabular}
    \caption{
    Dataset statistics and annotation granularity.
    A checkmark indicates that supervision is available at the corresponding level.
    }
    \label{tab:training-data}
\end{table}

We reserve 10.5K pages across IDL, TabMe++, and PubMed-OCR for evaluation
(Section~\ref{sec:in-domain}) and train on the remainder.
Held-out pages are selected uniformly at random over pages, stratified by
source and sequence length.
During each training stage, we additionally set aside 2{,}048 samples as a
validation set for early stopping; we halt training when the validation loss
saturates.
Appendix~\ref{app:data-splits} specifies the exact sampling and filtering
criteria.

\subsection{Training recipe}
\label{sec:training}

\paragraph{Base models.}
We fine-tune the public instruction-tuned Qwen2.5-VL-3B and Qwen2.5-VL-7B
checkpoints as backbones~\cite{bai2025qwen25vltechnicalreport}.
Both support long-context multimodal inputs and are comparable in scale to
recent document-focused VLMs, allowing us to study OCR performance as a
function of model size under a fixed training recipe.
During training, we rasterize each PDF page to a single 72 DPI image and 
feed it directly to the NaViT‑style Qwen2.5‑VL vision encoder without cropping or tiling (Appendix~\ref{app:setup}). 
Each training example consists of a single page plus a text prompt, 
with the total sequence length capped according to the stage‑specific limits in Table~\ref{tab:training-stages}.

\paragraph{Optimization and hardware.}
We fine‑tune using AdamW~\cite{loshchilov2019decoupledweightdecayregularization} 
on 8×H100 GPUs;
see Appendix~\ref{app:hyperparams} for full hyperparameters and implementation details.

\paragraph{Prompt variation and image references.}
All tasks are prompt-driven.
For each (\texttt{task}, \texttt{input}, \texttt{output}) tuple we maintain
a bank of prompt templates and sample one uniformly at training time.
We also randomize how the image is referenced by sampling from a small set of
determiners and nouns, e.g.,
\emph{``this/that/the provided/the attached''} $\times$
\emph{``document/image/picture/file/scan''}.
Thus prompts such as
\emph{``Where is \{text\} in this document?''},
\emph{``Find occurrences of \{text\} in the above image and return bounding boxes as JSON.''},
or
\emph{``Without returning the text, where are all the LINES in the attached scan?''}
are all seen during training.
We observed qualitatively more robust behavior under minor phrasing changes
at inference time, while leaving the underlying semantics unchanged.

\paragraph{Length-based curriculum.}
To stabilize optimization and handle complex layouts at long context, we adopt
a curriculum over total sequence length (image prompt plus full-page reading
output).
We begin with short sequences and gradually extend the maximum length while
shifting the dataset mixture, as summarized in
Table~\ref{tab:training-stages}.

\begin{table}[htb]
    \centering\small
    \begin{tabular}{c ccccc ccc p{4cm}}
    \toprule
    \textbf{Stage} & \multicolumn{5}{c}{\textbf{Datasets}} & \textbf{Length range} & \textbf{Samples} & \textbf{Steps} & 
    \textbf{Goal} \\ 
    & SDG & GL & IDL & T++ & PMD & & \\
    \midrule
    1  & \cmark & \cmark & \cmark & \cmark & \xmark & $<2$k          & 2.5M & 20K & Core training \\
    2  & \xmark & \xmark & \cmark & \cmark & \xmark & $[2$k, $8$k$]$ & 0.5M & 4K & Real specialization \\
    3a & \xmark & \xmark & \cmark & \cmark & \cmark & $[2$k, $8$k$]$ & 0.3M & 2K & PMD exposure \\
    3b & \xmark & \xmark & \xmark & \xmark & \cmark & $[8$k, $16$k$]$& 0.3M & 2K & PMD specialization \\
    \bottomrule
    \end{tabular}
    \caption{
    Training stages and dataset mixture.
    SDG = SynthDoG Grounding, GL = Grounded \LaTeX{}, IDL = OCR-IDL,
    T++ = TabMe++, PMD = PubMed-OCR.
    Samples are counted by number of pages seen by the model.
    }
    \label{tab:training-stages}
\end{table}

In Stage~1, we train on short sequences ($<2{,}048$ tokens) drawn from a mixture of synthetic and real sources: SynthDoG Grounding (SDG), Grounded \LaTeX{} (GL), OCR-IDL (IDL), and TabMe++ (T++). Stage~2 moves to medium-length sequences ($2{,}048$–$8{,}192$ tokens) and uses only real pages (IDL, TabMe++), dropping synthetic pages and plain page-level transcripts to emphasize structured JSON outputs. Stage~3a keeps the same length range, adds PubMed-OCR (PMD), and introduces paragraph-based reading to refine paragraph structure and long-form page handling. Stage~3b then focuses on the longest contexts ($8{,}192$–$16{,}384$ tokens) using only PMD, emphasizing multi-column scientific articles and very long transcripts. At each stage we sample uniformly over the task families in Section~\ref{sec:tasks} that are supported by the available annotations (for example, lines/paragraphs from OCR-IDL and TabMe++, paragraph-level reading and \texttt{text2d} from PubMed-OCR, and detection-only tasks from SynthDoG).

We refer to the resulting checkpoints as GutenOCR-{3B,7B} Stage~1--3b and evaluate all stages
under the grounded OCR metrics of Section~\ref{sec:metrics}.

\section{Evaluation Setup}

\subsection{Metrics}
\label{sec:metrics}

We evaluate grounded OCR along three axes:
(i) \emph{Text}---character- and word-level errors on string outputs;
(ii) \emph{Detection}---localization quality for boxes; and
(iii) \emph{End-to-end}---joint quality of box--text pairs and page-level fidelity
(including reading order and whitespace).
Table~\ref{tab:metrics-cheatsheet} summarizes which metrics are reported for
each task family.
Full formal definitions, normalization, and edge-case handling are deferred to
Appendix~\ref{app:metrics}.

\begin{table}[htb]
\centering\small
\begin{tabular}{lcc}
\toprule
\textbf{Task(s)} & \textbf{Output type} & \textbf{Reported metric(s)}\\
\midrule
Reading (\texttt{text}/\texttt{text2d}), Localized Reading
    & \texttt{text}
    & CER~$\downarrow$, WER~$\downarrow$\\
Detection, Conditional Detection
    & \texttt{boxes}
    & F1@0.5~$\uparrow$, Recall@0.5~$\uparrow$\\
Reading (\texttt{lines}/\texttt{paragraphs})
    & \texttt{text}+\texttt{boxes}
    & $\text{CER}_{\text{e2e}}$~$\downarrow$, mCER@0.5~$\downarrow$ \\
\bottomrule
\end{tabular}
\caption{
Metrics per task family at a glance.
Arrows indicate whether lower ($\downarrow$) or higher ($\uparrow$) values are better.
}
\label{tab:metrics-cheatsheet}
\end{table}

\paragraph{Character- and word-error rates.}
For any task that returns plain text, we report character error rate (CER)
and word error rate (WER).
Both are based on Levenshtein edit distance on lightly normalized strings
(Unicode NFKC + whitespace cleanup).
Intuitively, CER measures per-character mistakes, while WER is stricter and
counts word-level insertions, deletions, and substitutions.

\paragraph{Detection metrics.}
For detection-style outputs (plain or conditional), we report F1@0.5 and Recall@0.5.
Predicted and ground-truth boxes are matched one-to-one using IoU~$\ge 0.5$;
unmatched predictions count as false positives and unmatched ground-truth
boxes as false negatives.
The exact matching rule and treatment of corner cases (e.g., pages with no
boxes) are specified in Appendix~\ref{app:metrics}.

\paragraph{End-to-end structured reading.}
For structured reading tasks that produce \texttt{text}+\texttt{boxes}
(\texttt{lines}/\texttt{paragraphs}), we report two text metrics that reuse
the matched box set.
mCER@0.5 isolates recognition quality given successful localization:
we match predicted and ground-truth boxes at IoU~$\ge 0.5$, compute CER for
each matched text pair, and average over the matched set.
Unmatched ground-truth boxes do not affect mCER@0.5 directly (but do count as
false negatives in detection recall), cleanly separating “finding the right
region” from “reading it correctly.”

$\text{CER}_{\text{e2e}}$ measures page-level fidelity.
We linearize all predictions and references into page-level strings using a
deterministic reading order and a whitespace-preserving layout (as in the
\texttt{text2d} output format in Section~\ref{sec:tasks}), then compute CER
on these concatenated transcripts.
This metric is sensitive to missing or spurious lines, reading-order errors,
and layout-sensitive whitespace.

Together, these metrics probe (a) whether models find the right regions,
(b) whether they read those regions correctly, and (c) whether the resulting
page-level representation preserves reading order and layout-sensitive
whitespace.

For multi-task summaries (Section~\ref{sec:in-domain}), we use CER (or $\text{CER}_{\text{e2e}}$ for structured reading) as the reading error $\epsilon$, convert each into a per-task score $1-\epsilon$, and average these with detection and conditional-detection F1 values to obtain a composite grounded OCR score in $[0,1]$.

\subsection{Benchmarks}
\label{sec:benchmarks}

We evaluate GutenOCR on both in-domain business/scientific pages and public document benchmarks.

\paragraph{In-domain documents.}
We first evaluate on held-out pages drawn from the same sources used during
training (OCR-IDL, TabMe++, PubMed-OCR).
For each model we run the full suite of grounded OCR tasks from
Section~\ref{sec:tasks}: full-page reading in multiple output formats
(\texttt{text}, \texttt{text2d}, \texttt{lines}), localized reading given a
user-specified box, full-page detection, and conditional detection.
Metrics follow Table~\ref{tab:metrics-cheatsheet}: reading tasks are reported
as CER/WER errors, while detection tasks are summarized with F1@0.5 and
Recall@0.5; structured outputs use mCER@0.5 and $\text{CER}_{\text{e2e}}$ for
end-to-end quality.
Structured predictions are parsed through the shared JSON repair and
normalization pipeline in Appendix~\ref{app:parsing} before scoring.

\paragraph{Fox.}
Fox is a multi-page benchmark that probes a model's ability to ``focus
anywhere'' on a document via region-level, line-level, and color-guided OCR.
We follow the authors’ task definitions and restrict Fox to the four English
OCR-style tasks that align with our setting:
(i) page OCR, (ii) region-level OCR, (iii) line-level OCR, and
(iv) color-guided OCR.
For page OCR we report both the reading-order-agnostic soft token F1
(\emph{Page F1}) and the order-sensitive character error rate (\emph{Page CER});
for region, line, and color OCR we report CER only
(Table~\ref{tab:fox-bench}).
Additional benchmark details are given in Appendix~\ref{app:benchmarks}.

\paragraph{OmniDocBench v1.5.}
OmniDocBench v1.5 is a diverse PDF benchmark with dense annotations for layout
regions, text spans, formulas, and tables across multiple page types.
We treat it as an out-of-domain stress test and, restricting to English-only
pages, evaluate three behaviors that match our grounded OCR setting:
(i) text recognition via CER on cropped text spans,
(ii) formula recognition via CDM~\cite{wang2025imagetexttransformingformula}
and CER on cropped equations, and
(iii) text detection via recall of line-level text boxes.
Because OmniDocBench text spans cover only a subset of visible text, we report
recall@0.5 (but not precision/F1) for detection under our ``detect all readable
text'' ontology.
Full dataset and metric details are deferred to
Appendix~\ref{app:omnidoc}.

\section{Results}

We evaluate GutenOCR on three settings:
in-domain documents, Fox, and OmniDocBench v1.5. 
Our experiments are designed to answer three questions:
\begin{itemize}
    \item Does GutenOCR improve both reading and grounding over Qwen2.5-VL backbones?
    \item How does it compare to OCR-specialized and layout-aware models on standard benchmarks?
    \item How do curriculum stages trade off global reading quality vs fine-grained grounding?
\end{itemize}
We first describe the evaluated models (Section~\ref{sec:models}), 
then in-domain results (Section~\ref{sec:in-domain}), 
Fox (Section~\ref{sec:fox}),
OmniDocBench (Section~\ref{sec:omnidoc}), 
and a training-stage ablation (Section~\ref{sec:ablation}).

At a high level, GutenOCR-7B more than doubles the composite grounded OCR score of its Qwen2.5-VL-7B backbone on 10.5K held-out business and scientific pages (0.40$\rightarrow$0.82). On Fox, it substantially improves region- and line-level OCR while sacrificing some page-level linearization and color-guided OCR. On OmniDocBench v1.5, it raises text-detection recall from $\approx$0.02 to 0.55–0.62 but slightly degrades formula recognition, especially at 3B. We unpack these trends below.

\subsection{Models}
\label{sec:models}

We consider three groups of models (Table~\ref{tab:model-summary}): 
(i) general-purpose Qwen-based VLMs used via prompting,
(ii) OCR-specialized or layout-focused Qwen derivatives, and
(iii) our GutenOCR-3B/7B front-ends.
Full model descriptions are deferred to Appendix~\ref{app:models}.

\begin{table}[htb]
    \centering\small
    \begin{tabular}{l l l l c}
    \toprule
    \textbf{Model} & \textbf{Params} & \textbf{Backbone} & \textbf{Type / role} & \textbf{Coord.\ boxes} \\
    \midrule
    Qwen2.5-VL-3B/7B       & 3B / 7B   & -                 & Prompt-only VLM baseline     & \cmark \\
    Qwen3-VL-8B            & 8B        & -                 & Prompt-only VLM baseline     & \cmark \\
    Nanonets-OCR2-3B       & 3B        & Qwen2.5-VL-3B     & OCR-specialized baseline     & \xmark \\
    olmOCR 2               & 7B        & Qwen2.5-VL-7B     & OCR-specialized baseline     & \xmark \\
    Infinity-Parser-7B     & 7B        & Qwen2.5-VL-7B     & Layout parser baseline       & \xmark \\
    Chandra OCR            & 8B        & Qwen3-VL-8B       & Layout-preserving OCR        & \cmark \\
    DeepSeek-OCR           & 3B        & DeepSeek-3B-MoE   & OCR + grounding baseline     & \cmark \\
    PaddleOCR-VL-0.9B      & 0.9B      & PaddleOCR-VL-0.9B & OCR VLM (no layout model)    & \xmark \\
    GutenOCR               & 3B / 7B   & Qwen2.5-VL        & OCR + grounding (ours)       & \cmark \\
    \bottomrule
    \end{tabular}
    \caption{
    Overview of models considered in our evaluation.
    ``Coord.\ boxes'' indicates whether the public interface can emit bounding
    boxes or coordinates without additional tooling.
    }
    \label{tab:model-summary}
\end{table}

\subsection{In-Domain}
\label{sec:in-domain}

\begin{table}[htb]
    \centering\small
    \begin{tabular}{l cccc cc c}
        \toprule
        \multirow{2}{*}{\textbf{Model}} &
        \multicolumn{4}{c}{\textbf{Reading~$\downarrow$}} &
        \multicolumn{2}{c}{\textbf{Detection~$\uparrow$}} &
        \multirow{2}{*}{\textbf{Composite~$\uparrow$}} \\
        & \texttt{text} & \texttt{text2d} & \texttt{lines} & \texttt{local} &
          \texttt{full} & \texttt{conditional} & \\
        \midrule
        Qwen2.5-VL-3B  & 0.508 & 0.584 & 0.379 & 0.699 & 0.135 & 0.121 & 0.348 \\
        \quad Nanonets OCR2-3B & 0.369 & 0.518 & 0.645 & 0.807 & 0.001 & 0.001 & 0.277 \\
        \quad GutenOCR-3B (3a)
            & \underline{0.224}
            & \underline{0.293}
            & \underline{0.185}
            & \underline{\textbf{0.109}}
            & \underline{\textbf{0.798}}
            & \underline{0.877}
            & \underline{0.811} \\
        \cmidrule{2-8}
        Qwen2.5-VL-7B & 0.333 & 0.522 & 0.633 & 0.530 & 0.111 & 0.285 & 0.396 \\
        \quad Infinity-Parser-7B & 0.310 & 0.586 & 0.386 & 0.613 & 0.098 & 0.112 & 0.386 \\
        \quad olmOCR 2 & 0.365 & 0.530 & 0.515 & 0.745 & 0.039 & 0.024 & 0.318 \\
        \quad GutenOCR-7B (3a)
            & \underline{\textbf{0.202}}
            & \underline{\textbf{0.280}}
            & \underline{\textbf{0.147}}
            & \underline{0.129}
            & \underline{0.787}
            & \underline{\textbf{0.882}}
            & \underline{\textbf{0.819}} \\
        \cmidrule{2-8}
        Qwen3-VL-8B
            & \underline{0.285}
            & \underline{0.491}
            & \underline{0.306}
            & \underline{0.689}
            & \underline{0.049}
            & \underline{0.025}
            & \underline{0.384}
            \\
        \quad Chandra OCR
            & 0.390
            & 0.593
            & 0.867
            & 0.721
            & 0.012
            & 0.005
            & 0.241
            \\
        \cmidrule{2-8}
        DeepSeek-OCR
            & 0.702
            & 0.809
            & \underline{0.998}
            & 0.940 
            & \underline{0.010}
            & \underline{0.003}
            & 0.094 
            \\
        PaddleOCR-VL
            & \underline{0.399}
            & \underline{0.539}
            & 1.000
            & \underline{0.936}
            & 0.001
            & 0.002
            & \underline{0.188} \\
        \bottomrule
    \end{tabular}
    \caption{
        Average in-domain performance on OCR-IDL, TabMe++, and PubMed-OCR across six task families.
        Reading columns report errors (lower is better, $\downarrow$); detection columns report F1 scores (higher is better, $\uparrow$).
        The composite score averages task-wise values after mapping reading errors $\epsilon$ to scores $1-\epsilon$.
        Within each column, bold numbers indicate the best value across all models with a reported score, while underlined numbers indicate the best value within each backbone cohort (e.g., 3B vs.\ 7B).
        Entries that are both bold and underlined are best both overall and within their cohort; blank cells denote metrics that were not evaluated for that model.
    }
    \label{tab:in-domain-all}
\end{table}

We first assess in-domain performance on 10.5K held-out pages drawn from the
same real-world sources used in training (OCR-IDL, TabMe++, PubMed-OCR).
For each model (and, for GutenOCR, each training stage), we evaluate the six
task families from Section~\ref{sec:metrics}: full-page reading in three
output formats (\texttt{text}, \texttt{text2d}, \texttt{lines}), localized
reading, full-page detection, and conditional detection.
Table~\ref{tab:in-domain-all} reports CER for reading tasks and F1 for
detection and conditional detection.

To aggregate across tasks, we convert each reading error $\epsilon$ into a
score $1-\epsilon$ and then average these scores with the detection and
conditional-detection F1 values, yielding the composite column in
Table~\ref{tab:in-domain-all}.
Under this composite, GutenOCR-3B and GutenOCR-7B (Stage~3a) more than double
the scores of their Qwen2.5-VL backbones (from 0.348 to 0.811 and from 0.396
to 0.819, respectively), driven primarily by large gains in localized reading
and detection.
Unless otherwise noted, we use the Stage~3a checkpoints as the default
GutenOCR-3B and GutenOCR-7B models in subsequent experiments.

\subsection{Fox}
\label{sec:fox}

Recall from Section~\ref{sec:benchmarks} that we evaluate on the four English Fox OCR tasks that align with our training objectives: page OCR, region-level OCR, line-level OCR, and color-guided OCR (Table~\ref{tab:fox-bench}). Page-level performance is reported as both reading-order-agnostic token F1 (\emph{Page F1}) and order-sensitive CER, while the region, line, and color subtasks use CER only (lower is better for CER; higher is better for Page F1).

\begin{table}[htb]
    \centering\small
    \begin{tabular}{l ccccc}
    \toprule
    \textbf{Model} & \textbf{Page F1} $\uparrow$ & \textbf{Page} $\downarrow$ & \textbf{Region} $\downarrow$ & \textbf{Line} $\downarrow$ & \textbf{Color} $\downarrow$ \\
    \midrule
    Qwen2.5-VL-3B & 0.961 & 0.051 & 0.260 & 0.817 & 0.768 \\
    \quad Nanonets OCR2-3B & 0.976 & \underline{0.026} & 0.630 & 0.879 & \underline{0.475} \\
    \quad GutenOCR-3B & \underline{\textbf{0.988}} & 0.138 & \underline{\textbf{0.053}} & \underline{0.240} & 0.940 \\
    \cmidrule{2-6}
    Qwen2.5-VL-7B & \underline{0.984} & 0.025 & 0.163 & 0.701 & \underline{0.109} \\
    \quad olmOCR-2 & 0.983 & \underline{\textbf{0.018}} & 0.697 & 0.878 & 0.558 \\
    \quad Infinity-Parser-7B & 0.983 & 0.024 & 0.200 & 0.709 & 0.147 \\
    \quad GutenOCR-7B & 0.973 & 0.164 & \underline{0.067} & \underline{0.211} & 0.963 \\
    \cmidrule{2-6}
    Qwen3-VL-8B & \underline{0.971} & \underline{0.040} & \underline{0.468} & \underline{0.820} & \underline{0.200} \\
    \quad Chandra OCR & 0.954 & 0.053 & 0.794 & 0.958 & 0.852 \\
    \cmidrule{2-6}
    Fox$^\ast$ & -- & 0.046 & \underline{0.059} & \underline{\textbf{0.116}} & \underline{\textbf{0.064}} \\
    PaddleOCR-VL & \underline{0.975} & \underline{0.037} & 0.876 & 0.980 & 0.882 \\
    DeepSeek-OCR & 0.906 & 0.100 & 0.865 & 0.979 & 0.881 \\
    \bottomrule
    \end{tabular}
    \caption{
        Results on the English Fox OCR subtasks.
        \emph{Page F1} is reading-order agnostic soft token F1 for full-page OCR (higher is better);
        \emph{Page} is full-page OCR CER; \emph{Region} is region-level OCR given a
        provided box; \emph{Line} is line-level OCR given a line pointer; and
        \emph{Color} is OCR guided only by the color of a box.
        All entries except \emph{Page F1} are character error rates (CER; lower is better).
        Within each column, bold numbers indicate the best value across all models,
        while underlined numbers indicate the best value within each cohort
        (3B, 7B, 8B, and Fox/PaddleOCR-VL/DeepSeek-OCR, as separated by horizontal rules).
        Entries that are both bold and underlined are best both overall and within their cohort.
        Models with $^\ast$ are copied from the original Fox report; all others
        are recomputed under a unified prompting setup.
    }
    \label{tab:fox-bench}
\end{table}

\begin{figure}[htb!]
    \centering
    \begin{tabular}{cc}
        \begin{minipage}[t]{0.45\textwidth}
            \centering
            \includegraphics[width=0.8\textwidth]{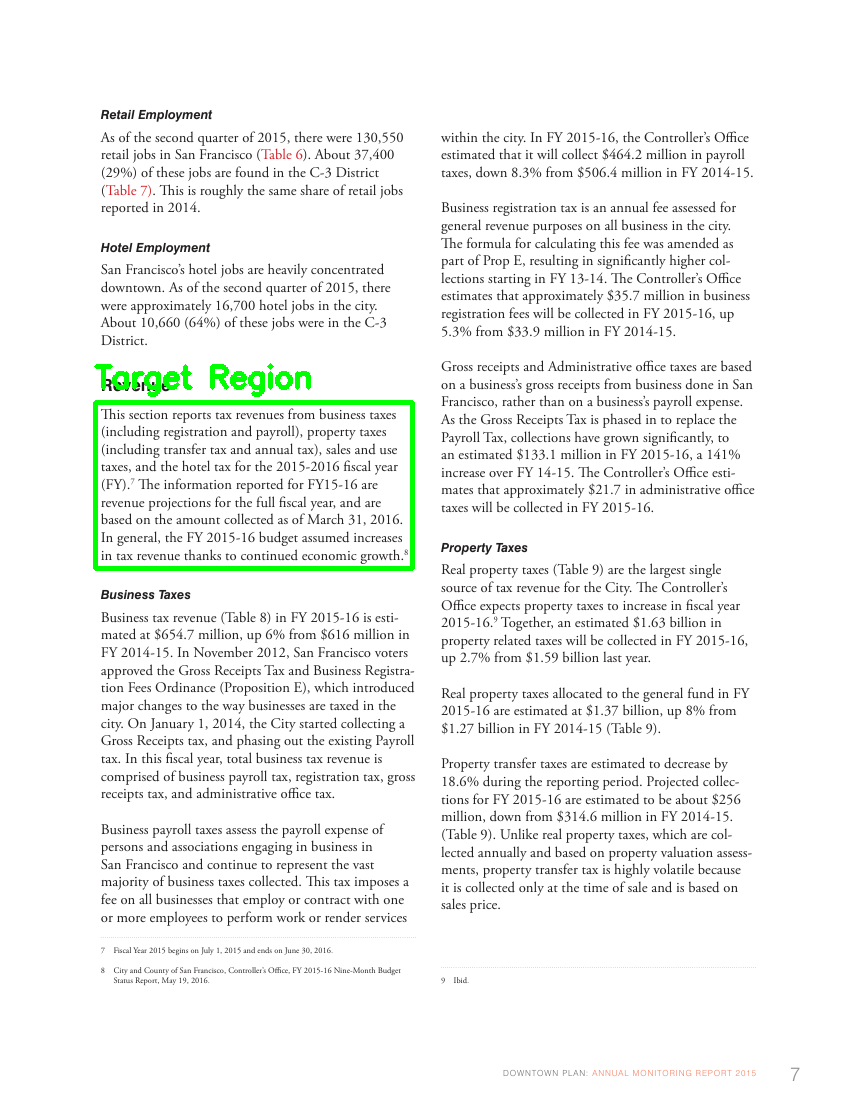}
            \\[0.3em]
            \scriptsize
            \textbf{GT:} \textit{This section reports tax revenues from busine...}
            \\
            \textcolor{red}{Qwen:} \textit{Revenue} (CER: 0.987)
            \\
            \textcolor{green}{Guten:} \textit{This section reports tax revenues from busine...} (CER: 0.004)
            \\[0.2em]
            \textbf{(a) Region-level OCR Success}
        \end{minipage}
        &
        \begin{minipage}[t]{0.45\textwidth}
            \centering
            \includegraphics[width=0.8\textwidth]{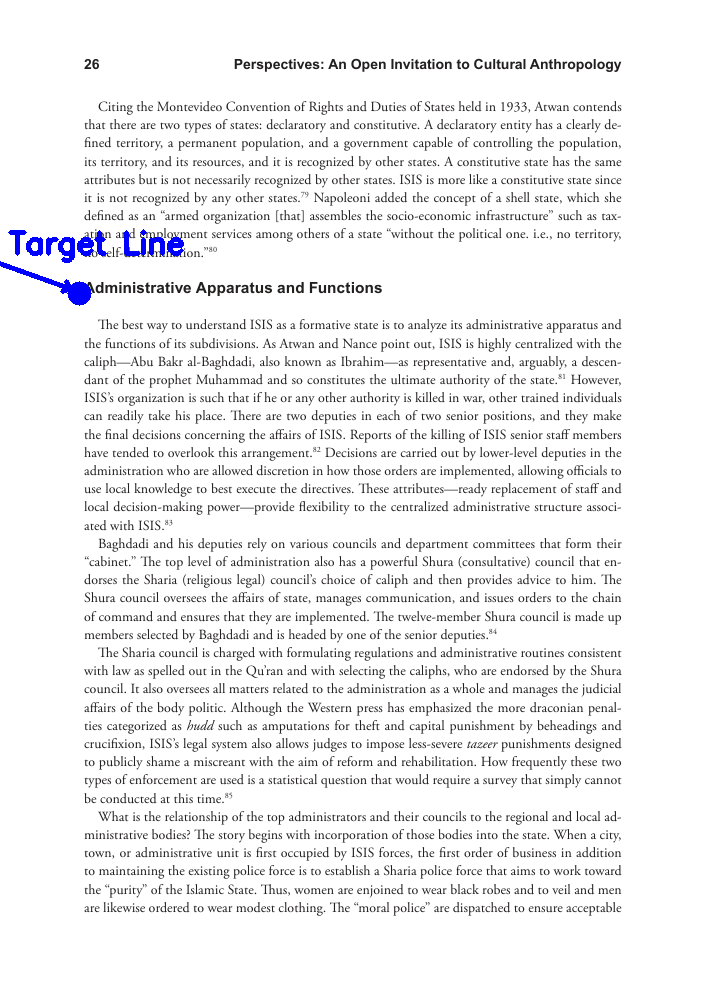}
            \\[0.3em]
            \scriptsize
            \textbf{GT:} \textit{Administrative Apparatus and Functions}
            \\
            \textcolor{red}{Qwen:} \textit{The best way to understand ISIS as a formativ...} (CER: 0.988)
            \\
            \textcolor{green}{Guten:} \textit{Administrative Apparatus and Functions} (CER: 0.000)
            \\[0.2em]
            \textbf{(b) Line Pointer Generalization}
        \end{minipage}
        \\[1em]
        \begin{minipage}[t]{0.45\textwidth}
            \centering
            \includegraphics[width=0.8\textwidth]{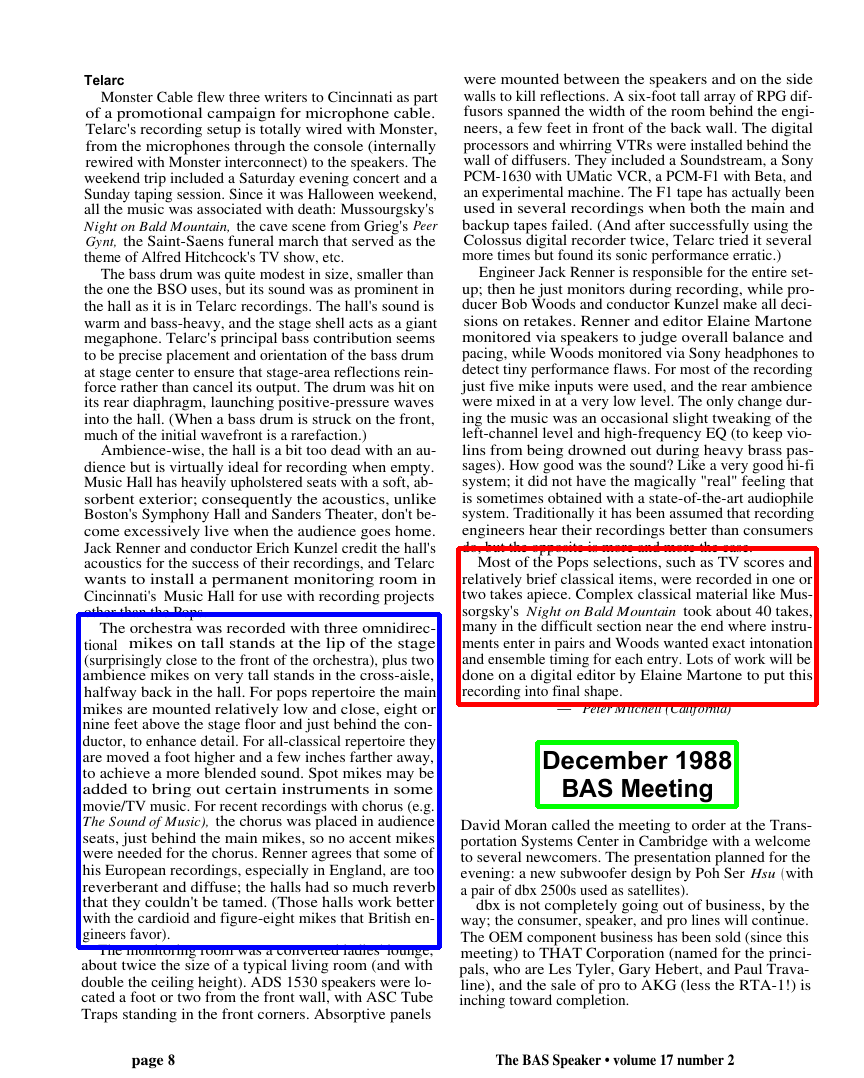}
            \\[0.3em]
            \scriptsize
            \textbf{GT:} \textit{December 1988 BAS Meeting}
            \\
            \textcolor{green}{Qwen:} \textit{December 1988 BAS Meeting} (CER: 0.000)
            \\
            \textcolor{red}{Guten:} \textit{[[461, 71, 809, 86], [461, 88, 809, 102], [84...} (CER: 0.997)
            \\[0.2em]
            \textbf{(c) Color OCR Catastrophic Forgetting}
        \end{minipage}
        &
        \begin{minipage}[t]{0.45\textwidth}
            \centering
            \includegraphics[width=0.8\textwidth]{media/en_15_page.png}
            \\[0.3em]
            \scriptsize
            \textbf{GT:} \textit{which could be undertaken for the development...}
            \\
            \textcolor{blue}{Qwen:} \textit{National advisory committees on human rights...}
            \\{\tiny (CER: 0.574, F1: 0.597)}
            \\
            \textcolor{purple}{Guten:} \textit{which could be undertaken for the development...}
            \\{\tiny (CER: 0.687, F1: 0.983)}
            \\[0.2em]
            \textbf{(d) Content vs Structure Trade-off}
        \end{minipage}
    \end{tabular}
    \caption{
        Qualitative comparison of Qwen2.5-VL-3B and GutenOCR-3B showing complementary strengths and weaknesses.
        \textbf{(a)} GutenOCR excels at region extraction (CER: 0.004 vs.\ 0.987).
        \textbf{(b)} Region training transfers to line pointer tasks (CER: 0.000 vs.\ 0.988).
        \textbf{(c)} GutenOCR suffers catastrophic forgetting on color tasks (CER: 0.997 vs.\ 0.000).
        \textbf{(d)} Trade-off: GutenOCR achieves higher content F1 (0.983 vs.\ 0.597) despite worse reading order (CER: 0.687 vs.\ 0.574) relative to Fox’s reference expectations.
    }
    \label{fig:fox-model-comparison}
\end{figure}

Fox mixes settings that are close to our training distribution (page- and
region-style OCR) with more extrapolative ones (line pointers and color-guided
grounding).
Page-level results highlight an important distinction between \emph{what}
GutenOCR reads and \emph{how} it linearizes that content.
At 3B, GutenOCR-3B attains the highest Page F1 (0.988), indicating that it reads almost all reference tokens, but its page-level CER is 0.138, much worse than Qwen2.5-VL-3B (0.051) and Nanonets-OCR2-3B (0.026). A similar pattern holds at 7B. The gap between Page F1 and Page CER reflects GutenOCR’s training objective: it is optimized for layout-sensitive 
\texttt{text2d} and structured line/paragraph JSON, not for Fox’s canonical page-to-Markdown order. Under Page F1 (content only), GutenOCR appears strong; Page CER (content + order) penalizes its layout-driven linearization.

This is consistent with our training objectives.
GutenOCR is optimized to produce layout-sensitive \texttt{text2d} outputs and
structured line/paragraph JSON, not a single fixed page-to-Markdown format.
The model is trained to follow 2D layout cues---columns, tables, headings, and
whitespace---when streaming text, which can conflict with the reading order
assumed by Fox’s reference markup.
Under Page F1 (which ignores order), this yields very high agreement on
content; under Page CER (which penalizes reordering and extra text), it appears
as relatively weak page OCR despite strong underlying reading ability.

On region-level OCR (\emph{Region}), GutenOCR is best aligned with the intent
of Fox.
GutenOCR-3B attains a region CER of 0.053, the best value in
Table~\ref{tab:fox-bench} and slightly better than the Fox model itself
(0.059), while GutenOCR-7B reaches 0.067.
Both represent large gains over the Qwen2.5-VL backbones (0.260 for 3B and
0.163 for 7B), consistent with our focus on grounded, region-aware OCR.

For line-level OCR (\emph{Line}), where Fox provides a pointer to a single line,
the Fox model remains strongest overall (0.116 CER), but GutenOCR again
substantially improves over its backbones.
GutenOCR-3B reduces line CER from 0.817 to 0.240, and GutenOCR-7B from
0.701 to 0.211.
Explicit line-pointer prompts do not appear in our training mixture, so this
behavior likely reflects transfer from region-level detection and localized
reading: once the model learns to find and read arbitrary regions, line
pointers become a natural special case.

Color-guided OCR (\emph{Color}) exposes a clear limitation of our current
recipe.
The Fox model, Qwen2.5-VL-7B, and several OCR-specialized systems leverage
color-coded boxes effectively (e.g., 0.064 CER for the Fox model and 0.109 CER
for Qwen2.5-VL-7B), whereas both GutenOCR variants perform poorly (0.940 CER
for 3B and 0.963 for 7B).
Color-conditioned pointers are absent from our training data, and the
OCR-focused fine-tuning overwrites the base models’ heuristic ability to
interpret these cues, a form of catastrophic forgetting on this narrow but
challenging sub-task.
Qualitatively, we often observe GutenOCR misreading the instruction as “find
the words \emph{red/green/blue}” rather than “read the text in the
red/green/blue box.”

GutenOCR explicitly trades some page-level CER on Fox for
stronger grounded behavior.
It preserves high Page F1 (indicating good reading) while reshaping reading
order to follow its layout-sensitive \texttt{text}/\texttt{text2d} training,
and it substantially strengthens region- and line-level OCR over the
Qwen2.5-VL backbones.
We qualitatively illustrate these trade-offs in
Figure~\ref{fig:fox-model-comparison} for the 3B-parameter model.

\subsection{OmniDocBench v1.5}
\label{sec:omnidoc}

Recall from Section~\ref{sec:benchmarks} that OmniDocBench v1.5~\cite{ouyang2025omnidocbenchbenchmarkingdiversepdf} provides dense annotations for layout regions, text spans, formulas, and tables across nine PDF page types (e.g., books, slides, financial reports, textbooks, exam papers, magazines, academic papers, handwritten notes, and newspapers).\footnote{See the v1.0$\rightarrow$1.5 changelog in the OmniDocBench supplement for details.}
The official benchmark is designed around end-to-end markup generation with structure-aware metrics, whereas our production stack targets JSON-style text+box outputs. Rather than retrofitting GutenOCR to OmniDocBench's canonical markup, we treat it as an out-of-domain stress test for three core capabilities:
\begin{enumerate}
    \item text recognition (CER on cropped text spans),
    \item formula recognition (CDM and CER on cropped equations), and
    \item text detection (recall of line-level text boxes).
\end{enumerate}
Throughout, we restrict evaluation to pages whose language tag is purely English.

\subsubsection{Text Recognition}

\begin{table*}[htb]
\centering
\small
\begin{tabular}{l cc ccc}
\toprule
\multirow{2}{*}{\textbf{Model}} &
\multicolumn{2}{c}{\textbf{Granularity}} &
\multicolumn{3}{c}{\textbf{Background}} \\
& Page & Sample & White & Single & Multi \\
\midrule
Qwen2.5-VL-3B  & 0.018 & 0.016 & 0.015 & 0.029 & 0.041 \\
GutenOCR-3B & 0.028 & 0.021 & 0.021 & 0.011 & 0.075 \\
\cmidrule{2-6}
Qwen2.5-VL-7B  & 0.011 & 0.010 & 0.010 & 0.014 & 0.026 \\
GutenOCR-7B & 0.024 & 0.020 & 0.020 & 0.012 & 0.037 \\
\bottomrule
\end{tabular}
\caption{
Component-level evaluation on the OmniDocBench OCR subset (English-only),
grouped by text attributes under the edit-distance metric.
All values are CER (lower is better).
}
\label{tab:omnidoc-ocr-components}
\end{table*}

For text recognition we use the OmniDocBench OCR subset, which provides cropped
text regions and attributes (language, background color, rotation).
We keep only English spans and evaluate character error rate (CER; lower is
better) under the same component-level edit-distance metric as the original
paper.
Following OmniDocBench, we report results at two granularities---full-page OCR
and per-sample crops---and further group spans by background type (white,
single-color, and multi-color).

Table~\ref{tab:omnidoc-ocr-components} summarizes results for the
Qwen2.5-VL backbones and their GutenOCR variants.
Across both model sizes, OCR-centric fine-tuning slightly increases CER
relative to the base models (e.g., from 0.018 to 0.028 page-level CER
for Qwen2.5-VL-3B, and from 0.011 to 0.024 for 7B).
The degradation is most pronounced on multi-colored backgrounds, where
CER more than doubles for both GutenOCR-3B and -7B, while the
Qwen2.5-VL baselines remain comparatively robust.

These results complement those from Fox (Section~\ref{sec:fox}), where
GutenOCR dramatically improves region- and line-level OCR but performs poorly
on color-guided pointers.
A plausible explanation is domain mismatch: the OmniDocBench OCR subset
contains many visually diverse pages (magazines, notes, newspapers,
slides) that are under-represented in our training corpus, which focuses
on business documents and structured forms, with academic pages added
only in the final training phase.
Color and background cues are also not modeled explicitly in our current
recipe, so the fine-tuning does not inherit the base models’ background
robustness on this benchmark.

\begin{table}[htb]
\centering
\small
\begin{tabular}{lcccccccc}
\toprule
Models & Book & Slides & Textbook & Exam & Magazine & Academic & Notes & Newspaper \\
\midrule
Qwen2.5-VL-3B & 0.014 & 0.007 & 0.047 & 0.025 & 0.007 & 0.024 & 0.000 & 0.018\\
GutenOCR-3B & 0.034 & 0.023 & 0.039 & 0.077 & 0.006 & 0.018 & 0.000 & 0.008 \\
\cmidrule{2-9}
Qwen2.5-VL-7B & 0.011 & 0.008 & 0.029 & 0.016 & 0.007 & 0.009 & 0.167 & 0.005 \\
GutenOCR-7B & 0.034 & 0.027 & 0.031 & 0.057 & 0.004 & 0.014 & 0.000 & 0.004 \\
\bottomrule
\end{tabular}
\caption{
OmniDocBench OCR subset: CER by page type (English-only; lower is better).
}
\label{tab:omnidocbench-ocr-domains}
\end{table}

\begin{figure}[htb]
  \centering
  \includegraphics[width=0.7\linewidth]{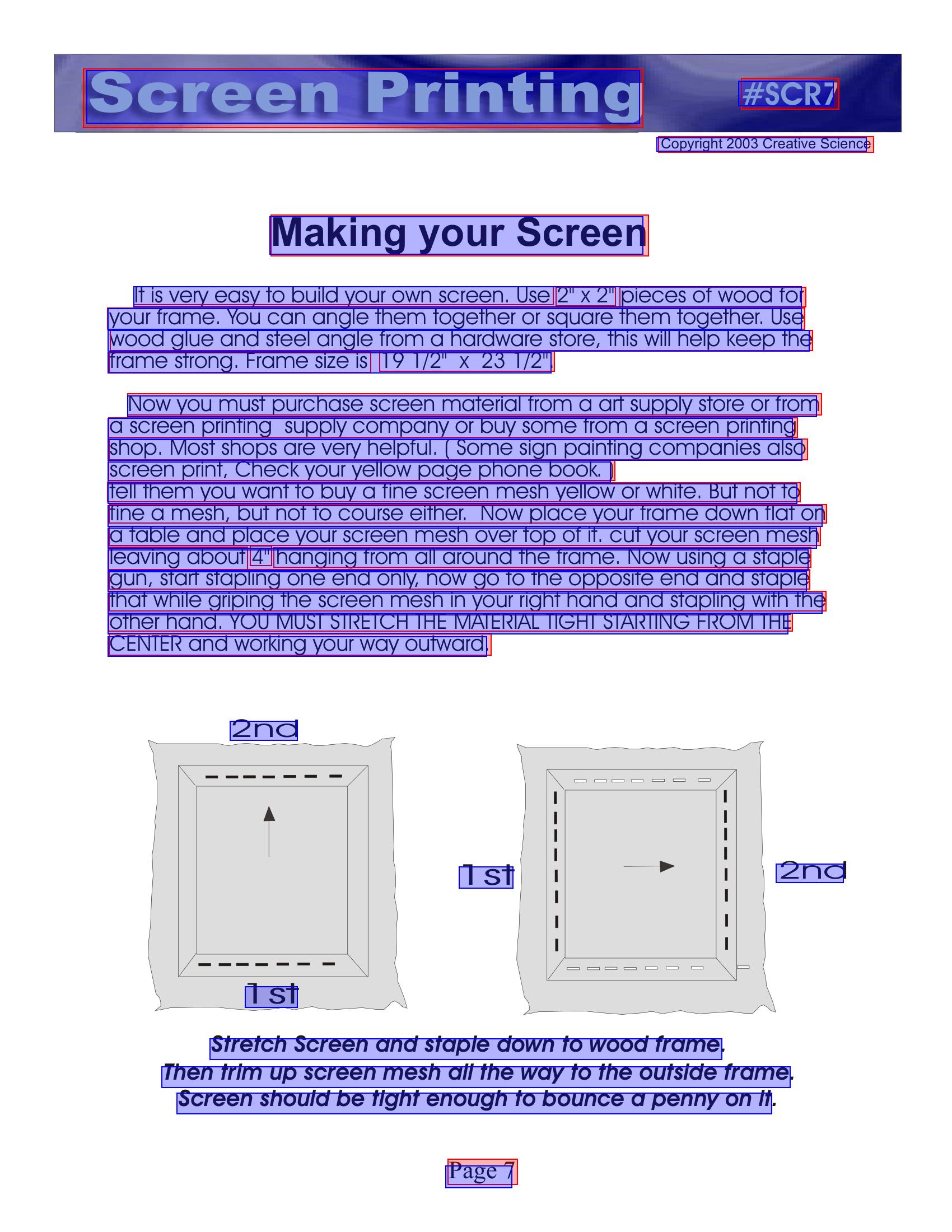}
  \caption{
  Qualitative OmniDocBench text-detection example.
  Red boxes denote annotated text spans; blue boxes denote GutenOCR predictions
  on the same page.
  Many blue boxes fall on readable text (e.g., table cells and decorative
  labels) that OmniDocBench does not label as text, so non-overlapping
  predictions cannot be reliably treated as false positives.
  This motivates our recall-only evaluation protocol in
  Section~\ref{sec:omnidoc-text-detection}.
  }
  \label{fig:omnidoc-text-det-example}
\end{figure}

\subsubsection{Text Detection}
\label{sec:omnidoc-text-detection}

For text detection, several design choices in OmniDocBench interact with our
modeling goals.
Text spans are defined at roughly line level and are provided only for a subset
of visible text: many visually text-like objects (e.g., table cell content,
certain chemical formulas, decorative labels) are either annotated only as part
of a \texttt{Table} object or tagged as ``Equation Ignore'' and excluded from
text evaluation.
GutenOCR, by contrast, aims to detect \emph{all} readable text, including text
inside tables and equations, and does not attempt to reproduce
OmniDocBench’s span-selection heuristics.
As illustrated in Figure~\ref{fig:omnidoc-text-det-example}, our predictions
(blue) cover substantially more text-like regions than the red annotated spans.

Because OmniDocBench’s text annotations cover only a subset of visible text (for example, many table cells and some equations are unlabeled), we cannot reliably treat non-overlapping predictions as false positives. A predicted box that misses an annotated span may be a hallucination or may simply hit unannotated but readable text. We therefore evaluate only recall at IoU~$\geq 0.5$, treating the annotated spans as a trusted but incomplete ground truth and asking: when OmniDocBench says there is text here, does our detector cover it?

\begin{table}[htb]
\centering
\small
\begin{tabular}{l c cccccccc}
\toprule
Model & All & Book & Slides & Textbook & Exam & Magazine & Academic & Notes & Newspaper \\
\midrule
Qwen2.5-VL-3B & 0.022 & 0.028 & 0.005 & 0.055 & 0.038 & 0.041 & 0.004 & 0.000 & 0.038 \\
GutenOCR-3B & 0.620 & 0.679 & 0.707 & 0.758 & 0.471 & 0.797 & 0.493 & 0.000 & 0.773 \\
\cmidrule{2-10}
Qwen2.5-VL-7B & 0.020 & 0.045 & 0.005 & 0.012 & 0.015 & 0.015 & 0.014 & 0.000 & 0.035 \\
GutenOCR-7B & 0.549 & 0.645 & 0.648 & 0.639 & 0.503 & 0.705 & 0.399 & 0.000 & 0.759 \\
\bottomrule
\end{tabular}
\caption{
OmniDocBench layout subset: text detection recall at IoU~$\geq$~0.5
(R@0.5) by page type (higher is better).
}
\label{tab:omnidoc-det}
\end{table}

Under this recall-only protocol (Table~\ref{tab:omnidoc-det}), the Qwen2.5-VL
backbones almost never produce boxes that align with OmniDocBench’s text spans
(R@0.5~$\approx 0.02$ for both 3B and 7B).
After GutenOCR fine-tuning, recall increases by more than an order of
magnitude, reaching 0.620 for GutenOCR-3B and 0.549 for GutenOCR-7B.
This is consistent with our design: the GutenOCR curriculum explicitly trains a
line-level detection head on diverse business documents, whereas the
original Qwen2.5-VL models were never optimized to emit dense text boxes that
match an external annotation standard.

We emphasize that these numbers should not be interpreted as full detection
quality on OmniDocBench: higher recall here does not imply stronger precision
on their ontology, and our models deliberately predict many additional boxes
(e.g., table cells, formulas) that the benchmark does not label as text.
Rather, OmniDocBench serves as an external check that the detection behavior
learned on our in-domain datasets transfers reasonably well to a visually
diverse, attribute-rich document collection.
Accordingly, we view OmniDocBench text detection as a recall-only stress test rather than a full detector ranking: higher R@0.5 indicates better coverage of annotated spans, but does not reflect precision or behavior on unannotated text.

\subsubsection{Formula Recognition}

\begin{table}[htb]
    \centering
    \small
    \begin{tabular}{l cc ccccc}
    \toprule
    & & \multicolumn{6}{c}{\textbf{CDM}~$\uparrow$} \\
    \textbf{Model} & \textbf{CER}~$\downarrow$ & Overall & Book & Slides & Textbook & Exam & Academic \\
    \midrule
    Qwen2.5-VL-3B & 0.189 & 0.936 & 0.951 & 0.876 & 0.979 & 0.958 & 0.914 \\
    GutenOCR-3B & 0.294 & 0.866 & 0.881 & 0.758 & 0.975 & 0.907 & 0.864 \\
    \cmidrule{2-8}
    Qwen2.5-VL-7B  & 0.216 & 0.935 & 0.951 & 0.877 & 0.991 & 0.955 & 0.914 \\ 
    GutenOCR-7B & 0.221 & 0.927 & 0.940 & 0.870 & 0.906 & 0.951 & 0.915 \\
    \bottomrule
    \end{tabular}
    \caption{
    Formula recognition evaluation on the English subset of OmniDocBench.
    CER is lower-is-better; CDM is higher-is-better.
    }
    \label{tab:omnidoc-formula-recognition}
\end{table}

OmniDocBench includes a dedicated formula subset with LaTeX ground truth and
evaluates models with the Character Detection Matching (CDM)
metric~\cite{wang2025imagetexttransformingformula}, which measures symbol-level
correctness under rendering, together with normalized edit distance.
We follow this setup, again filtering to English formulas, and report CDM
(higher is better) together with CER in
Table~\ref{tab:omnidoc-formula-recognition}.

On formulas, GutenOCR exhibits clear negative transfer relative to
its Qwen2.5-VL bases.
For the 3B model, CDM drops from 0.936 to 0.866 and CER rises from 0.189
to 0.294; for the 7B model, the degradation is milder but consistent:
CDM decreases from 0.935 to 0.927 and CER increases from 0.216 to 0.221.

These results suggest that the current GutenOCR training mixture is effectively
math-agnostic (likely exacerbated by dropping LaTeX grounding early on): it
substantially improves document OCR
(Sections~\ref{sec:in-domain}--\ref{sec:fox}) but erodes formula recognition,
especially for the 3B model.
For equation-heavy use cases, the best-performing checkpoints in our
experiments are the un-fine-tuned Qwen2.5-VL backbones, not the GutenOCR
variants.
Making GutenOCR explicitly math-aware---for example, by incorporating
formula-rich supervision similar to the OmniDocBench formula subset---is an
important direction for future work.

\subsection{Training Stage Ablation}
\label{sec:ablation}

\begin{figure}[htb]
  \centering
  \includegraphics[width=0.45\linewidth]{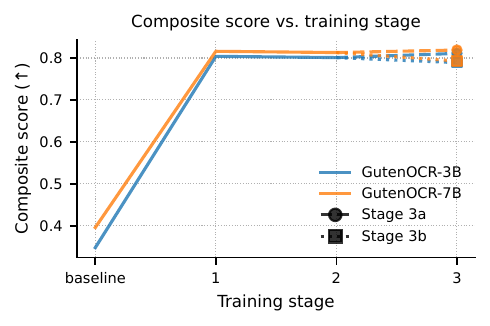}
  \hfill
  \includegraphics[width=0.45\linewidth]{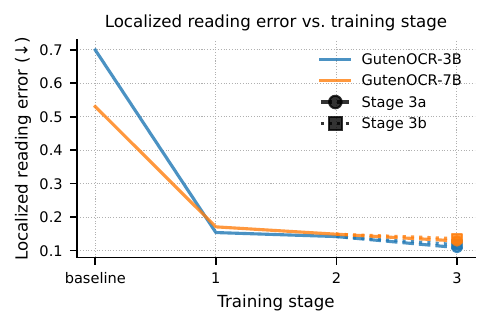}
  \caption{
    Training-stage ablation summary for composite score (left, higher is better)
    and localized-reading error (right, lower is better) on OCR-IDL, TabMe++,
    and PubMed-OCR.
    The x-axis shows the base model and Stages~1--3; Stage~3a and Stage~3b
    share the same x-position and are distinguished by line style and marker.
  }
  \label{fig:training-ablation}
\end{figure}

To study curriculum trade-offs, we compare GutenOCR checkpoints across
training stages on the same held-out in-domain pages.
Table~\ref{tab:in-domain-eval} reports averages over OCR-IDL, TabMe++, and
PubMed-OCR for each stage, and
Figure~\ref{fig:training-ablation} summarizes the corresponding trends in
composite score and localized-reading error.

\begin{table}[htb]
    \centering
    \small
    \begin{tabular}{lc ccc c c c c}
    \toprule
    \multirow{ 2}{*}{\textbf{Model}} & \multirow{ 2}{*}{\textbf{Stage}} & \multicolumn{4}{c}{\textbf{Reading~$\downarrow$}} & \multicolumn{2}{c}{\textbf{Detection~$\uparrow$}} & \multirow{ 2}{*}{\textbf{Composite~$\uparrow$}} \\ 
    & & \texttt{text} & \texttt{text2d} & \texttt{lines} & \texttt{local} & \texttt{full} & \texttt{conditional} & \\
    \midrule
    Qwen2.5-VL-3B  &    & 0.508 & 0.584 & 0.379 & 0.699 & 0.135 & 0.121 & 0.348 \\
    \quad GutenOCR-3B & 1  & \underline{0.218} & \underline{0.276} & \underline{0.183} & 0.154 & \underline{\textbf{0.799}} & 0.854 & 0.804 \\
    \quad GutenOCR-3B & 2  & 0.223 & 0.282 & 0.185 & 0.142 & 0.788 & 0.851 & 0.801 \\
    \quad GutenOCR-3B & 3a & 0.224 & 0.293 & 0.185 & \underline{\textbf{0.109}} & 0.798 & \underline{0.877} & \underline{0.811} \\
    \quad GutenOCR-3B & 3b & 0.253 & 0.345 & 0.208 & 0.118 & 0.790 & 0.868 & 0.789 \\
    \cmidrule{3-9}
    Qwen2.5-VL-7B  &    & 0.333 & 0.522 & 0.633 & 0.530 & 0.111 & 0.285 & 0.396 \\
    \quad GutenOCR-7B & 1  & \underline{\textbf{0.198}} & \underline{\textbf{0.255}} & \underline{\textbf{0.141}} & 0.171 & \underline{0.790} & 0.871 & 0.816 \\
    \quad GutenOCR-7B & 2  & 0.210 & 0.263 & 0.154 & 0.149 & 0.778 & 0.876 & 0.813 \\
    \quad GutenOCR-7B & 3a & 0.202 & 0.280 & 0.147 & \underline{0.129} & 0.787 & \underline{\textbf{0.882}} & \underline{\textbf{0.819}} \\
    \quad GutenOCR-7B & 3b & 0.242 & 0.323 & 0.169 & 0.136 & 0.761 & 0.866 & 0.793 \\
    \bottomrule
    \end{tabular}
    \caption{
        Average in-domain performance on OCR-IDL, TabMe++, and PubMed-OCR
        across six task families, broken down by training stage.
        Reading columns report errors (lower is better, $\downarrow$); detection
        columns report F1 scores (higher is better, $\uparrow$).
        The composite score averages task-wise values after mapping reading
        errors $\epsilon$ to scores $1-\epsilon$.
        Within each column, bold numbers indicate the best value across all
        models and stages, while underlined numbers indicate the best value
        within each backbone (3B vs.\ 7B).
        Entries that are both bold and underlined are best both overall and
        within their backbone.
    }
    \label{tab:in-domain-eval}
\end{table}

Relative to the Qwen2.5-VL backbones, GutenOCR yields large and consistent
gains across all in-domain tasks.
For the 3B model, the best stage (3a) roughly halves full-page reading errors
(e.g., \texttt{text} CER 0.508$\rightarrow$0.224), cuts localized-reading error
from 0.699 to 0.109, and lifts detection and conditional-detection F1 from
0.135/0.121 to $\sim$0.8/0.88, more than doubling the composite
(0.348$\rightarrow$0.811).
The 7B backbone shows the same pattern: GutenOCR-7B (3a) reduces reading errors
by 40--80\%, raises detection F1 from 0.111 to 0.787 and conditional detection
from 0.285 to 0.882, and increases the composite from 0.396 to 0.819.
Thus most of the in-domain improvement comes from the OCR-centric training
recipe, not from scaling alone.

After fine-tuning, model-size differences are modest.
At their best stages (3a), GutenOCR-7B improves the composite only slightly
over GutenOCR-3B (0.819 vs.\ 0.811).
The 7B model consistently achieves lower errors on global reading tasks
(\texttt{text}, \texttt{text2d}, \texttt{lines}) and slightly higher overall
composite, while the 3B model achieves lower localized-reading error at every
stage and slightly higher unconditional detection F1.
This trade-off suggests that GutenOCR-3B is a competitive, more efficient
choice when local reading and pure detection are prioritized, whereas
GutenOCR-7B is preferable when global reading and conditional detection are
the main objectives.

Within each backbone, the curriculum stages exhibit clear trends.
Stage~1 already captures most of the gains in full-page reading and detection relative to the base VLMs. Stage~2 makes only minor adjustments and is largely dominated by Stage~1 or Stage~3a.
Adding Stage~3a mainly sharpens localized reading and conditional detection for
both backbones and yields the best composite scores, indicating that the
additional curriculum primarily refines grounding rather than further reducing
global reading error.
By contrast, Stage~3b slightly degrades the composite relative to Stage~3a and
does not produce any column-best metrics, consistent with mild
over-specialization on the longest PubMed-OCR-style contexts.

Overall, Table~\ref{tab:in-domain-eval} and
Figure~\ref{fig:training-ablation} show that GutenOCR turns general-purpose
VLMs into strong, grounded OCR systems across in-domain documents, with small
but consistent advantages for the 7B backbone and clear stage-wise
curriculum trade-offs.

\subsection{Qualitative Examples}

\begin{figure}
    \centering
    \begin{tabular}{cc}
        \begin{minipage}[t]{0.45\textwidth}
            \centering
            \includegraphics[width=\textwidth]{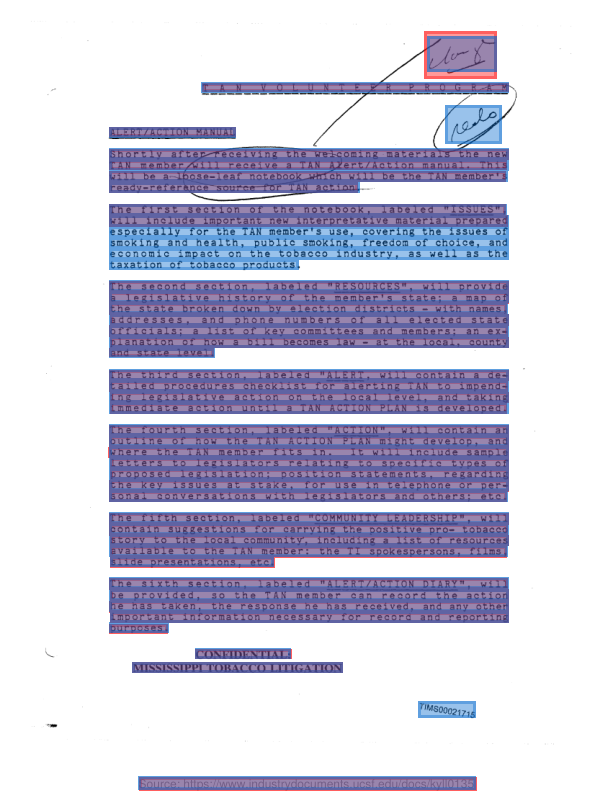}
            \\[0.3em]
            \scriptsize
            \textbf{(a) IDL}
        \end{minipage}
        &
        \begin{minipage}[t]{0.45\textwidth}
            \centering
            \includegraphics[width=\textwidth]{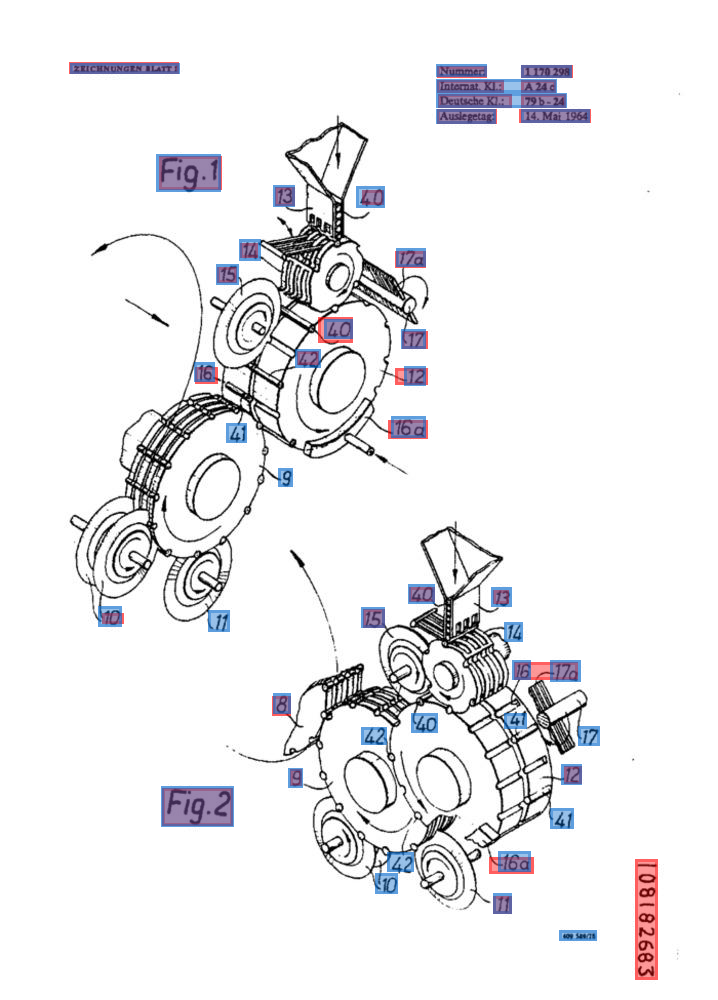}
            \\[0.2em]
            \scriptsize
            \textbf{(b) TabMe++}
        \end{minipage}
        \\
        \begin{minipage}[t]{0.35\textwidth}
            \centering
            \includegraphics[width=\textwidth]{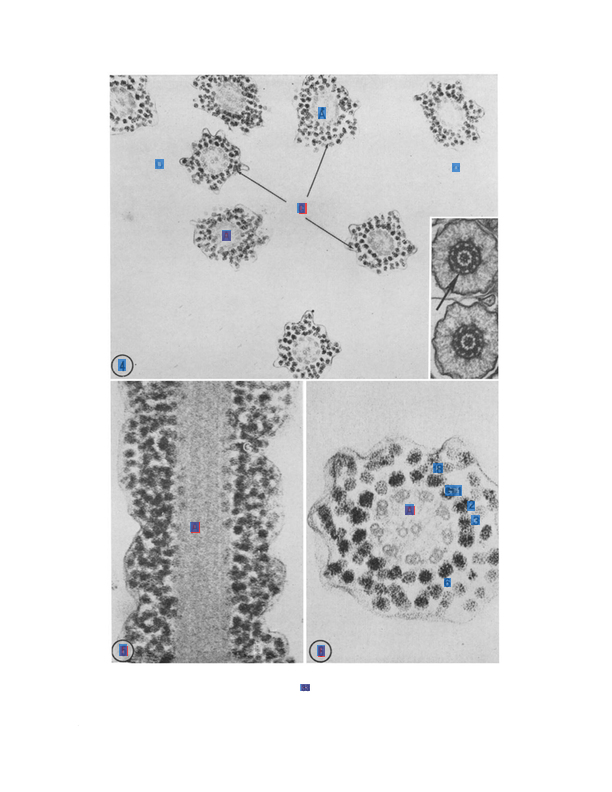}
            \\[0.2em]
            \scriptsize
            \textbf{(c) PubMed-OCR}
        \end{minipage}
        & 
        \begin{minipage}[t]{0.3\textwidth}
            \centering
            \includegraphics[width=\textwidth]{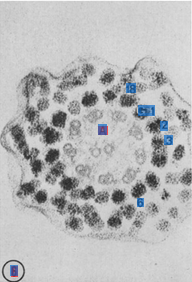}
            \\[0.2em]
            \scriptsize
            \textbf{(d) PubMed-OCR (zoom-in)}
        \end{minipage}
        \\
    \end{tabular}
    \caption{Qualitative examples of GutenOCR-3B's \texttt{lines} predictions on test samples. Blue boxes indicate GutenOCR's prediction versus the ground truth in red. Regions of overlap are purple. Areas of disagreement are immediately obvious (and actionable).}
    \label{fig:qualitative}
\end{figure}

To exemplify how GutenOCR can be used to further improve annotations
with a human-in-the-loop, we highlight an example on each real dataset
where the 3B parameter model correctly found \emph{more} lines than were
originally annotated by the ground truth OCR engine in Figure~\ref{fig:qualitative}.

\section{Discussion}
\label{sec:discussion}

Taken together, our results suggest that treating OCR as a grounded
front-end---rather than as a single page-to-markup conversion---changes
both what should be optimized and how OCR outputs are consumed in
practice.
In particular, when OCR is used as the substrate for downstream
retrieval, extraction, and document QA, the most costly failures are
often not minor transcription noise but missing regions, unstable
reading order, and weak provenance between model outputs and the pixels
that support them.

\paragraph{Grounded front-ends versus page-only OCR.}
Across our in-domain suite, Fox, and OmniDocBench, GutenOCR behaves
differently from both its Qwen2.5-VL backbones and contemporary
OCR-specialized VLMs.
Rather than optimizing a single notion of page-level fidelity, our
recipe emphasizes a \emph{grounded front-end}: high recall of text
regions, stable line- and paragraph-level outputs, and predictable
behavior under localized and conditional reading.
This shows up quantitatively in large gains on localized reading and
detection metrics (Tables~\ref{tab:in-domain-all} \& \ref{tab:fox-bench}),
and qualitatively in cases where GutenOCR recovers legible content that
is absent from reference annotations.
From a systems perspective, these properties are often as important as
small differences in page-level CER: downstream components can tolerate
some transcription noise, but they are brittle to missing content,
scrambled layout, or non-reproducible pointing behavior that prevents
targeted re-reading.

\paragraph{Human-in-the-loop verification and control.}
Viewed through the lens of human review, GutenOCR is designed to reduce
the effort required to decide whether a page is \emph{usable} for
downstream automation.
Line- and paragraph-level detection, together with multi-format
transcripts (plain text, layout-preserving text, and line-based JSON),
let a reviewer quickly answer concrete questions:
(i) \emph{Did we cover all the text?} (recall of boxes over the page),
(ii) \emph{Can we reliably re-read a region of interest?} (localized
reading that is stable across prompts), and
(iii) \emph{Can we attribute extracted values to evidence?} (boxes that
pinpoint the supporting span).
This asymmetric cost profile motivates our bias toward recall and
fine-grained grounding: false positives are typically easy to dismiss
visually, whereas false negatives force re-scanning the page and make
exceptions difficult to triage.
In practice, these properties support review tools, exception
workflows, and active-learning loops where the goal is not only
accurate text but \emph{quickly verifiable} text and \emph{actionable
provenance}.
By contrast, page-to-Markdown systems often entangle rendering choices
with OCR itself, making isolated errors or ambiguous attributions
harder to localize without repeatedly comparing a long transcript to
the original page.

\paragraph{Interfaces, semantic surface, and comparison to prior work.}
A large portion of recent work binds OCR to a single, task-specific
output representation: page-to-Markdown/HTML conversion, fixed
layout-to-JSON schemas, or document parsing pipelines that produce one
canonical ``document object'' for downstream use.
These approaches can be strong end-to-end baselines when the downstream
task matches the target representation, but they often provide limited
support for (i) alternative reading orders, (ii) region-restricted
re-reading, and (iii) fine-grained attribution of answers to the
specific spans that support them.
In particular, document conversion systems (e.g., those centered on
structured markup such as DocTags-like formats~\cite{nassar2025smoldoclingultracompactvisionlanguagemodel}) provide exportable
parses that are valuable for rendering and indexing, but they do not by
themselves guarantee \emph{evidence-first} behavior---that is, a stable
contract where extracted fields and QA answers can be traced to precise
token- or span-level locations on the page.

GutenOCR takes a different position in this design space.
It offers a small set of stable, grounded interfaces---full-page
reading in multiple formats, localized reading, full-page detection,
and conditional detection---all backed by a single checkpoint and a
coherent prompting scheme.
These interfaces act as a thin, API-like layer between raw pixels and
downstream logic: applications consume structured outputs with text and
boxes and are free to render or transform them into whatever schema
they require (HTML, Markdown, XML, database records) without changing
the underlying OCR model.
We view this as exposing a broader \emph{semantic surface}---a set of
stable OCR primitives that downstream systems can reliably compose---than
page-only transcripts, while remaining predictable enough for
enterprise integration and human verification.

\paragraph{Limitations and open challenges.}
Our experiments also reveal the limits of the current recipe.
First, specialization for business and scientific documents improves
reading and grounding on our in-domain suite and on Fox’s region and
line tasks, but can reduce robustness on visually diverse layouts and
color-guided cues, and degrades formula recognition on formula-heavy
OmniDocBench pages.
Second, our outputs are still primarily text- and line-centric: we do
not yet explicitly model table structure, math layout, cross-page
links, or higher discourse units, even though these are critical for
many real workflows (e.g., attributing a financial value to the correct
row/column context).
Addressing these weaknesses will likely require both a broader training
mixture (e.g., math- and table-rich supervision, color- and
background-aware tasks) and extended interfaces that surface this
structure \emph{without sacrificing} grounding and controllability.
Finally, because our goal is a practical front-end, future work should
also characterize efficiency trade-offs (latency/throughput) alongside
accuracy and grounding.

\paragraph{Toward document holograms.}
The preceding points suggest a target beyond OCR alone.
We use the term \emph{document hologram} to describe a persistent,
multi-view representation of a document that couples (i) fine-grained
grounding between content and pixels (down to spans or tokens), with
(ii) learned semantic typing and relations over that grounded content,
and (iii) queryable views that support region-restricted inspection and
controllable reading orders.
Crucially, a document hologram is not merely an export format or parse
tree: it is the internal state that enables \emph{evidence-bearing}
responses, where extracted fields and QA answers come with a set of
supporting locations (often many-to-many) that a user can directly
verify---for example, pointing to the specific cells in a large table
that justify a suspected fraudulent amount.

In this sense, GutenOCR should be viewed as a front-end module for
document holograms rather than a complete solution.
Our results show that a general-purpose VLM can be specialized into a
grounded OCR front-end that substantially improves structured reading
and detection while preserving localized and pointer-based behaviors
useful for evidence retrieval.
At the same time, the observed failure modes---math and color
sensitivity, table-heavy layouts, and cross-page reasoning---highlight
what is still missing from higher hologram layers that model structure
and semantics on top of grounded text.
We leave the design of these higher layers, and the evaluation
protocols that directly test evidential QA and fine-grained provenance,
to future work.

\section{Related Work}

\subsection{Classical and Modern OCR Pipelines}

Classical OCR systems decompose document understanding into a sequence of stages: image normalization and
binarization, text region detection, character- or word-level recognition, and downstream post-processing. Early
systems such as Tesseract~\cite{TessOverview} and successors in industrial settings (e.g., PaddleOCR~\cite{cui2025paddleocr30technicalreport}
and cloud APIs such as Amazon Textract~\cite{amazon_textractor} and Google Vision OCR~\cite{google_cloud_ocr})
largely share this modular structure, even as individual components have shifted from rule-based processing to
convolutional detectors and sequence recognizers. Production deployments typically add task-specific logic for
page segmentation, tables, forms, and key-value extraction.

Despite their maturity, pipeline OCR systems exhibit well-known limitations on complex documents.
First, layout handling can be brittle: multi-column articles, nested tables, stamps, marginalia, and rotated inserts can
break reading-order heuristics, scrambling text streams and propagating errors downstream~\cite{wangLayoutReaderPretrainingText2021,zhangReadingOrderMatters2023}.
Second, the link between text and visual support is often indirect: while many systems expose word- or line-level geometry,
programmatic control over reading behavior (alternative reading orders, region-restricted rereading, whitespace-sensitive
linearizations, or query-conditioned extraction) typically requires substantial glue logic on top of OCR outputs~\cite{karmanovEclairExtractingContent2025}.
Finally, hybrid document pipelines frequently combine PDF-native extraction (when embedded text is available) with OCR
fallback for scanned or rasterized content; recent work on PDF-centric QA highlights the value of exploiting PDF structure
and metadata when available~\cite{saadfalcon2023pdftriagequestionansweringlong}.

In this work, we adopt the pipeline formulation as our starting point: we aim to preserve the strengths of modular OCR---
explicit detection, recognition, and structured outputs---while replacing hand-crafted components with a unified
vision--language model. Our GutenOCR system is trained to perform full-page reading, detection, conditional detection,
and localized reading within a single VLM interface, behaving like a traditional OCR pipeline from the outside while enabling
fine-grained grounding, controllable reading order, and task-specific document representations.

\subsection{Document Conversion Toolchains and Markup Formats}

A complementary line of work targets \emph{document conversion}: converting PDFs and page images into richly structured,
LLM-friendly intermediate representations such as Markdown, HTML, or JSON. Docling provides an open-source conversion
toolkit that parses multiple document formats into a unified structured representation~\cite{livathinosDoclingEfficientOpenSource2025}.
Recent compact conversion models such as SmolDocling generate \emph{DocTags}, an XML-like tagged format designed
to preserve document structure and element locations while remaining tokenizer-friendly~\cite{nassar2025smoldoclingultracompactvisionlanguagemodel}.
IBM's Granite-Docling-258M similarly targets efficient end-to-end conversion into structured machine-readable formats
with tight integration into the Docling toolchain~\cite{ibm_granite_docling_258m_2025}.
Other open-source tools (e.g., MinerU) focus on practical PDF-to-Markdown/JSON conversion for scientific and technical
documents~\cite{wang2024mineruopensourcesolutionprecise,niu2025mineru25decoupledvisionlanguagemodel}.

These systems provide strong end-to-end baselines when a downstream application can consume their target schema
directly. Our goal is different: rather than committing to a single canonical export format, we emphasize a small set of
\emph{OCR primitives as a stable contract} (detection, reading, localized reading, and conditional detection) that downstream
systems can compose into diverse workflows and representations.

\subsection{Vision--Language Models for Documents}

Recent document understanding systems increasingly rely on VLMs that operate on page images.
One family consists of \emph{OCR-based} models such as LayoutLM and successors~\cite{Xu_2020,xu-etal-2021-layoutlmv2,10.1145/3503161.3548112},
DocFormer~\cite{Appalaraju_2021_ICCV}, DiT~\cite{10.1145/3503161.3547911}, LiLT~\cite{wang2022lilt}, and UDOP~\cite{tang2023udop},
which consume OCR tokens and bounding boxes. More recently, layout-aware LLM extensions such as DocLLM~\cite{wang-etal-2024-docllm}
and DocLayLLM~\cite{liaoDocLayLLMEfficientMultimodal} incorporate spatial structure directly into generative LMs, but still assume OCR has produced
accurate text and geometry. These approaches achieve strong results on downstream benchmarks (e.g., FUNSD~\cite{jaume2019},
CORD~\cite{park2019cord}, SROIE~\cite{li2019icdar2019}, DocVQA~\cite{Mathew2020DocVQAAD}), but are limited by upstream OCR errors and provide
only indirect control over region-restricted reading or query-conditioned localization.

A second family of \emph{OCR-free} document VLMs operates directly on pixels and emits text or structured markup.
Donut~\cite{kim2022donut} is a prominent example; related work on screenshot/document parsing (e.g., Pix2Struct~\cite{lee2023pix2structscreenshotparsingpretraining})
and scientific document conversion (e.g., Nougat~\cite{blecher2023nougatneuralopticalunderstanding}) likewise targets page-to-markup generation. Several newer OCR-specialized
parsers (e.g., dots.ocr~\cite{li2025dotsocrmultilingualdocumentlayout}, DocPedia~\cite{fengDocPediaUnleashingPower2024a}) emphasize faithful conversion and reading order, often producing
Markdown/HTML/JSON (sometimes with coordinates) and evaluating primarily via text/markup edit distance.

More recently, several OCR-specialized multimodal models fine-tune VLM backbones for high-fidelity document
transcription and parsing. General OCR Theory (GOT) frames OCR as a unified end-to-end ``OCR-2.0'' model that
supports whole-page transcription as well as interactive, region-guided recognition (e.g., via coordinates or colors) and
formatted outputs~\cite{wei2024generalocrtheoryocr20}.
Other systems use reinforcement learning or verifiable reward signals to improve structural fidelity and reading order in
page-to-text/markup conversion, including olmOCR~\cite{poznanski2025olmocr2unittest} and Infinity-Parser~\cite{wang2025infinityparserlayoutaware}.
These models provide strong end-to-end conversion baselines, but they typically expose a primary canonical output
representation (plain text or markup) rather than a small set of schema-stable OCR primitives spanning pure detection,
localized re-reading, and string-conditioned search.

In parallel, general-purpose vision foundation models (e.g., Florence-2~\cite{xiao2023florence2advancingunifiedrepresentation}, Granite Vision~\cite{granitevisionteam2025granitevisionlightweightopensource},
Qwen2.5-VL/Qwen3-VL~\cite{bai2025qwen25vltechnicalreport,bai2025qwen3vltechnicalreport}, InternVL variants~\cite{chen2024internvl,zhu2025internvl3,wang2025internvl3_5})
can perform document QA, captioning, and OCR-like behaviors when suitably prompted. However, their interfaces and training objectives are
typically optimized for broad multimodal instruction following rather than a well-specified OCR API contract, and coordinate outputs are
often exposed through ad-hoc formats.

Our work takes a different stance: we explicitly retrain a general-purpose VLM backbone as a grounded OCR front-end with task schemas that
mirror a classical OCR pipeline (full-page reading, detection, localized reading, and conditional detection). The primary product is a set of
structured text+box outputs and prompt-selectable behaviors, rather than QA answers or a single canonical document linearization.

\subsection{Grounded OCR and Localization}

Before the rise of VLMs, grounded OCR was studied through \emph{text spotting} systems that jointly detect and recognize text
(e.g., EAST~\cite{Zhou_2017_CVPR}, CTPN~\cite{tian2016detecting}, FOTS~\cite{Liu_2018_CVPR}, Mask TextSpotter~\cite{Lyu_2018_ECCV}).
These models provide explicit box--text pairs, but are often tuned for scene text and do not naturally support document abstractions or
language-driven queries.

More recently, work on reading order and layout-aware extraction (e.g., LayoutReader~\cite{wangLayoutReaderPretrainingText2021},
Reading Order Matters~\cite{zhangReadingOrderMatters2023}, \'{e}clair~\cite{karmanovEclairExtractingContent2025}) assumes OCR tokens exist.
Meanwhile, phrase-grounding models (e.g., MDETR~\cite{kamath2021mdetrmodulateddetection}, GLIP~\cite{li2021grounded,zhang2022glipv2},
Grounding DINO~\cite{liu2024groundingdinomarryingdino,ren2024groundingdino15advance}) align language with regions; document-focused tasks
appear in benchmarks such as Fox~\cite{liu2024fox}. Foundation models such as Florence-2 expose OCR and region-level prompting as part of a
unified multi-task interface~\cite{xiao2023florence2advancingunifiedrepresentation}.

Several OCR-focused VLM parsers annotate outputs with coordinates or support prompted localization (e.g., dots.ocr~\cite{li2025dotsocrmultilingualdocumentlayout},
SmolDocling/DocTags~\cite{nassar2025smoldoclingultracompactvisionlanguagemodel}). However, these interfaces are typically optimized for end-to-end conversion rather than a
single checkpoint that unifies \emph{pure detection}, \emph{localized rereading}, and \emph{string-conditioned search} under a stable,
schema-driven API with standardized conditional-localization metrics.

GutenOCR can be viewed as a generalization of text spotting and phrase grounding to a unified, VLM-based OCR interface: one checkpoint supports
(i) full-page detection, (ii) full-page structured reading, (iii) localized reading given a user box, and (iv) conditional detection given a query
string, all expressed through prompt-selectable schemas.

\subsection{Benchmarks and Evaluation for OCR and Document Understanding}
\label{sec:related-benchmarks}

Classical OCR evaluation has been driven by ICDAR-style benchmarks for scene text and document text recognition, which typically separate detection and recognition. Detection is measured with IoU-based precision, recall, and F1 at one or more thresholds; recognition quality is reported as character error rate (CER), word accuracy, or normalized edit distance on isolated words, lines, or cropped regions. End-to-end text spotting benchmarks often combine localization and recognition through matching-based protocols
(e.g., end-to-end F-score/Hmean), and more fine-grained schemes such as CLEval score character-level correctness while
handling split/merge cases in detection~\cite{baek2020cleval}.

Document understanding benchmarks broaden the focus from text recognition to classification and extraction. Datasets such as RVL-CDIP~\cite{harley2015icdar}, FUNSD~\cite{jaume2019}, CORD~\cite{park2019cord}, SROIE~\cite{li2019icdar2019}, and DocVQA~\cite{Mathew2020DocVQAAD} evaluate document image classification, form understanding, key-value extraction, and question answering, typically assuming access to an external OCR system. In these settings, OCR quality is a hidden variable: models are judged on downstream task accuracy, and errors stemming from misrecognition, mis-localization, or poor reading order are entangled.

More recent benchmarks have started to evaluate document parsing and grounding more directly. Fox~\cite{liu2024fox} introduces multi-page, fine-grained tasks such as region-level OCR, line-level OCR, and color-guided OCR, using CER on cropped regions as the primary metric. OmniDocBench v1.5~\cite{ouyang2025omnidocbenchbenchmarkingdiversepdf} provides dense annotations for layout regions, text spans, formulas, and tables across a diverse set of page types; it reports component-level edit distance for text, CDM and edit distance for formulas~\cite{wang2025imagetexttransformingformula}, and structure-aware metrics such as TEDS for tables~\cite{zhong2020imagebasedtablerecognitiondata}. These benchmarks substantially raise the bar on the diversity and richness of evaluation. At the same time, they are
primarily organized around either component-wise parsing fidelity (text/formula/table/reading-order modules) or
end-to-end markup correctness under a particular canonicalization~\cite{ouyang2025omnidocbenchbenchmarkingdiversepdf}.
They do not directly evaluate interactive OCR primitives such as region-restricted re-reading or string-conditioned
conditional detection under a unified, schema-driven contract.
Recent OCR benchmarks for large multimodal models also evaluate OCR across multiple task tracks (e.g., multilingual
reading, document parsing, and key information extraction), highlighting common failure modes such as grounding
errors and repetition hallucinations~\cite{yang2024ccocrcomprehensivechallengingocr,fu2025ocrbenchv2improvedbenchmark}.

Our evaluation protocol builds on these traditions but targets \emph{grounded OCR} as a first-class objective. Following Section~\ref{sec:metrics}, we measure: (i) text accuracy via a balanced CER and WER on whole-page transcripts and localized crops; (ii) detection quality via F1@0.5 and Recall@0.5 over axis-aligned boxes; and (iii) end-to-end quality via CERe2e and mCER@0.5 on aligned box--text pairs, which together disentangle localization errors (missing, spurious, or mis-ordered boxes) from recognition errors within correctly localized regions. Unlike prior work that reports either text-only metrics or IoU-only metrics, this metric set evaluates (a) whether models find the right regions, (b) whether they read those regions correctly, and (c) whether the resulting page-level representation preserves reading order and layout-sensitive whitespace. In this sense, our metrics are an explicit evolution of classical OCR and text-spotting criteria toward the grounded, layout-aware document representations that document-hologram systems require.

\section{Conclusion}
GutenOCR recasts OCR as a grounded front-end problem, training a single vision--language checkpoint to support full-page reading, detection, localized reading, and conditional detection through a unified prompt interface. Across in-domain business and scientific documents, Fox, and OmniDocBench v1.5, it substantially improves structured reading and detection over Qwen2.5-VL backbones and several OCR-specialized baselines, while revealing clear failure modes in formula-heavy and color-guided settings. These results suggest that explicit grounding and controllable reading are as important as page-level transcripts for downstream document systems and point toward richer “document hologram” representations that we leave for future work.

\bibliographystyle{unsrt}  
\bibliography{references}  

\appendix

\section{Interface contract}
\label{app:interface}

\subsection{Task families \& I/O schemas}
\label{app:interface-tasks}

For each task family in Section~\ref{sec:tasks}, we parameterize the interface
by a \emph{task type} (\texttt{reading}, \texttt{detection},
\texttt{conditional\_detection}, \texttt{localized\_reading}),
an \emph{input type} (image only, image + query string, image + bounding box),
and an \emph{output type} (\texttt{text}, \texttt{text2d},
\texttt{lines}, \texttt{paragraphs}, \texttt{BOX}).

At training time we maintain a small bank of prompt templates for each
(task, input, output) triple and sample one uniformly, to encourage robustness
to minor paraphrases.
Below we list representative templates for each family; in all cases
\texttt{\{FORMAT\}} is replaced with one of the output types
(\texttt{text}, \texttt{text2d}, \texttt{lines}, \texttt{paragraphs}):

\paragraph{Full-page reading.}
\begin{itemize}
  \item ``Read all text in the provided document and return the result as \{FORMAT\}.''
  \item ``Transcribe the text from this page. Use the \{FORMAT\} format described above.''
  \item ``Without adding explanations, output only the \{FORMAT\} representation of the page.''
\end{itemize}

\paragraph{Full-page detection.}
\begin{itemize}
  \item ``Without returning any text, detect all LINES in this document and output their bounding boxes as a JSON array.''
  \item ``Locate all PARAGRAPHS in the image and return a JSON array of boxes \([x_1,y_1,x_2,y_2]\). Do not include text.''
  \item ``Detect all math regions in the page and return their bounding boxes as JSON.''
\end{itemize}

\paragraph{Conditional detection.}
\begin{itemize}
  \item ``Where does the exact string \texttt{"\{q\}"} appear in this document? Return a JSON array of bounding boxes for all matching lines.''
  \item ``Search for \texttt{"\{q\}"} in the page and output \([x_1,y_1,x_2,y_2]\) boxes for each line that contains it. If there are no matches, return \texttt{[]}.''  
\end{itemize}

\paragraph{Localized reading.}
\begin{itemize}
  \item ``What text appears inside the region \([x_1,y_1,x_2,y_2]\) of this page? Return only the recognized text.''
  \item ``Read the text inside the specified box and output a \texttt{lines} or \texttt{paragraphs} string, without JSON or extra commentary.''
\end{itemize}

During training, we also randomize the noun used to refer to the image
(e.g., ``document'', ``image'', ``page'', ``scan'') and simple determiners
(e.g., ``this'', ``that'', ``the attached'', ``the provided'').
We qualitatively found that this light prompt augmentation improves robustness to
prompt phrasing at inference time while keeping the semantics of the
task fixed.

\subsection{Geometry and \texttt{BOX} Type}
\label{app:interface-geometry}

All grounded outputs ultimately reduce to axis-aligned bounding boxes in the
coordinate system of the input image.

\paragraph{Coordinate system.}
We use integer pixel coordinates with origin at the top-left corner of the
rasterized page.
A bounding box is represented as
\[
[x_1, y_1, x_2, y_2]
\]
with \(x_1 < x_2\) and \(y_1 < y_2\).
Here \(x\) increases to the right and \(y\) increases downward.

\paragraph{Axis-aligned boxes only.}
We do not predict rotated boxes.
Rotated or skewed text is represented by the axis-aligned bounding box of its
minimal enclosing rectangle.
This choice simplifies evaluation and downstream consumption.

\paragraph{Clipping and degeneracies.}
During post-processing we apply the following normalizations to both
predictions and references:
\begin{itemize}
  \item Clip all coordinates to the closed interval \([0, W]\times[0, H]\),
        where \(W\) and \(H\) are the image width and height in pixels.
  \item Discard degenerate boxes with non-positive width or height
        (i.e., \(x_1 \ge x_2\) or \(y_1 \ge y_2\)).
\end{itemize}
These steps ensure that all boxes used for evaluation have strictly positive
area and lie within the image bounds.

\paragraph{Grounded output schema.}
Whenever an output type is grounded and includes text, we use JSON arrays of
objects of the form
\begin{verbatim}
{"text": <string>, "bbox": [x1, y1, x2, y2]}
\end{verbatim}
For pure detection tasks, where no text is required, the output is a JSON
array of bounding boxes only:
\begin{verbatim}
[
  [x1, y1, x2, y2],
  [x1, y1, x2, y2],
  ...
]
\end{verbatim}
We apply a tolerant JSON-repair step before evaluation and treat irreparable
outputs as invalid; see Appendix~\ref{app:parsing} for details.

\subsection{Layout-Sensitive \texttt{text2d}}
\label{app:interface-text2d}

The \texttt{text2d} representation is designed to encode 2D layout using only
spaces and newline characters while remaining a plain string.
It is constructed from line-level grounded outputs of the form
\texttt{\{"text","bbox"\}} as follows.

\paragraph{Preprocessing.}
We begin with a set of line predictions
\(\{(\hat{t}_i, b_i)\}_{i=1}^N\), where
\(b_i = [x_{1,i}, y_{1,i}, x_{2,i}, y_{2,i}]\).
We discard boxes with non-positive width or height and clip all remaining
boxes to the image bounds.
We then compute simple global statistics (e.g., median line height and
a crude character density in characters per pixel) that are used to
approximate how many fixed-width ``character cells'' fit in a typical line.

\paragraph{Line ordering.}
We sort lines according to a 2D reading order: first by the vertical position
of the line center (top-to-bottom), then by the horizontal position of the
left edge (left-to-right) within each row.
This ordering approximates a natural reading order across multi-column
layouts and figures.

\paragraph{Mapping to a notional grid.}
We conceptually render lines into a notional character grid.
For each ordered line, we:
\begin{itemize}
  \item Map the absolute horizontal position of its left edge to a character
        column index using the estimated character density.
  \item Insert spaces to account for horizontal gaps between consecutive lines
        on the same row (e.g., to right-justify page numbers or align columns).
  \item Emit the normalized line text \(\norm{\hat{t}_i}\) at the
        corresponding columns.
\end{itemize}
Vertical gaps between consecutive lines are measured in pixels using the
distance between their vertical centers.
If this gap exceeds a threshold proportional to the estimated line height, we
insert one or more blank lines in the output.
In practice we cap the number of inserted blank lines to a small maximum to
avoid pathological vertical expansion.

\paragraph{Final string.}
After all lines have been placed, we obtain a 2D grid with characters and
spaces.
We trim trailing spaces on each row and convert the grid to a single string
by joining rows with newline characters.
The result is a \texttt{text2d} string that:
\begin{itemize}
  \item preserves relative horizontal alignment via runs of spaces,
  \item preserves vertical structure via explicit blank lines, and
  \item can be consumed by downstream systems as a plain-text API.
\end{itemize}

During evaluation, we preserve whitespace in \texttt{text2d} 
so that layout-sensitive differences (e.g., scrambled columns or collapsed
vertical gaps) are reflected in the error.

\subsection{Task-Specific Semantics}
\label{app:interface-semantics}

\paragraph{Conditional detection.}
Each conditional-detection example consists of a page image, a query string
\(q\), and a target set of line-level boxes.
To construct the target, we take the reference line transcripts and apply
the same normalization $\norm{\cdot}$ used for text metrics
(Appendix~\ref{app:metrics}), including Unicode NFKC normalization, trimming
of leading/trailing whitespace, and collapsing of internal whitespace runs.
We apply $\norm{\cdot}$ to both the query and each line transcript and mark a
line as a hit if the normalized query appears as a case-sensitive substring
of the normalized line.

If the query does not appear on the page, the ground truth set is empty and
the expected model output is an empty JSON array
\texttt{[]} with no additional commentary.
When a query appears multiple times on the same line, we treat that as a
single target box; if it appears on multiple lines, all corresponding line
boxes are included in the target set.
During evaluation we compare predicted and reference boxes with IoU-based
matching at a threshold of 0.5; unmatched predicted boxes are false positives
and unmatched reference boxes are false negatives.

\paragraph{Localized reading.}
For localized reading, the input consists of a page image and a bounding box
$b = [x_1,y_1,x_2,y_2]$ in pixel coordinates.
The model is expected to transcribe only the text inside $b$.
We train both line-level and paragraph-level variants, but in both cases the
\emph{output} is treated as non-grounded: the model returns a single plain-text
string rather than a JSON array of \texttt{\{"text","bbox"\}} objects.

To construct training labels, we intersect $b$ with the reference
line- or paragraph-level boxes and concatenate the transcripts of blocks whose
overlap with $b$ exceeds an IoU threshold (we use 0.5 in all experiments).
The resulting string is used as the target transcript for that localized
reading example.
At evaluation time, we compute CER and WER between the model output and this
target string, again using $\norm{\cdot}$ to normalize whitespace and
Unicode variants.

\paragraph{Full-page reading and detection.}
For full-page reading tasks, the model receives only the page image and is
prompted to produce either:
\begin{itemize}
  \item a linear transcript (\texttt{text}),
  \item a whitespace-preserving transcript (\texttt{text2d}), or
  \item a structured array of line-/paragraph-level objects
        (\texttt{lines}/\texttt{paragraphs}).
\end{itemize}
Structured predictions are parsed through the JSON repair and normalization
pipeline described in Appendix~\ref{app:parsing}.
Invalid or irreparable outputs are treated as failures and assigned maximal
error according to Appendix~\ref{app:metrics}.

For full-page detection, the model receives only the page image and must
output a JSON array of bounding boxes that cover all lines or paragraphs of
interest, without transcripts.
Supervision and metrics follow the detection protocol in
Appendix~\ref{app:metrics}, using IoU-based matching at threshold 0.5.

\section{Data}
\label{app:data}

\subsection{Real corpora: filtering and normalization}
\label{app:data-real}

For OCR-IDL and PubMed-OCR we start from document-level PDFs and
associated OCR outputs provided by each source.
We apply a common preprocessing pipeline to obtain page images and normalized
ground-truth annotations.

All PDFs are rasterized to single-page RGB images at 72\,dpi.
TabMe++ was rasterized in the initial construction of TabMe and is used as is.

\paragraph{Text normalization.}
For each available transcript (paragraph-, line-, or word-level) we apply the
same normalization operator $\norm{\cdot}$ used for text metrics
(Appendix~\ref{app:metrics}), including:
\begin{itemize}
  \item Unicode NFKC normalization,
  \item conversion of platform-specific quotes, dashes, and bullets to
        canonical ASCII or Unicode codepoints,
  \item trimming of leading/trailing whitespace, and
  \item collapsing of internal whitespace runs into a single space.
\end{itemize}
This produces consistent supervision across datasets and aligns training-time
targets with evaluation-time normalization.

\paragraph{Geometry unification.}
All datasets provide boxes in pixel coordinates, but differ in conventions
(e.g., half-open vs.\ closed intervals, one-based vs.\ zero-based indices).
We map all coordinates to the axis-aligned \([x_1,y_1,x_2,y_2]\) convention
from Section~\ref{sec:tasks}, clip to image bounds, and discard degenerate
boxes with non-positive width or height.

\subsection{Synthetic corpora: generation pipelines}
\label{app:data-synth}

\paragraph{Grounded \LaTeX{} (GL).}
The GL corpus is designed to provide highly localized supervision for
mathematical expressions.

\begin{enumerate}
  \item \textbf{Equation mining.}
        We extract raw \LaTeX{} fragments from Wikimedia dumps following
        \cite{schubotz2025wiki}.
  \item \textbf{Rendering.}
        For each equation we create a blank page and render the equation at a
        randomly sampled:
        \begin{itemize}
          \item rotation angle, and
          \item sub-pixel position on the page.
        \end{itemize}
\end{enumerate}

\paragraph{SynthDoG grounding (SDG).}
The SDG corpus adapts SynthDoG~\cite{kim2022donut} to better match our
line-centric grounding objectives.

\begin{enumerate}
  \item \textbf{Text source.}
        We draw raw text snippets from FinePDFs~\cite{kydlicek2025finepdfs},
        sampling both narrative and transactional content.
        Snippets are truncated to at most a few hundred characters to keep
        synthetic pages compact.
  \item \textbf{Layout generation.}
        We use SynthDoG's layout engine to place text blocks, but modify the
        line-breaking logic to be word-aware: line breaks are only inserted at
        whitespace, and hyphenation is disabled.
        This avoids mid-word splits that complicate line-level supervision.
  \item \textbf{Line-level supervision.}
        We instrument the generator to emit line-level bounding boxes and
        transcripts directly, in addition to any word-level boxes that
        SynthDoG already produces.
        Paragraph-level groupings are not used in SDG.
  \item \textbf{Rendering and noise.}
        Pages are rendered at 300\,dpi with randomized fonts, font sizes, and
        background textures (e.g., paper grain).
        We apply the same noise augmentations as in GL (blur, compression,
        low-amplitude geometric jitter).
\end{enumerate}

\subsection{Splits and held-out sets}
\label{app:data-splits}

\paragraph{Evaluation pages.}
We reserve $\sim$10{,}500 pages across IDL, TabMe++, and PubMed-OCR for
evaluation (Section~\ref{sec:in-domain}).
Pages are sampled uniformly at random from the union of these three corpora.
All remaining pages from IDL, TabMe++, and PubMed-OCR are used for training
or validation.

\paragraph{Training and validation splits.}
Within each training stage we randomly carve out 2{,}048 examples from the
current training pool as a validation set.
Sampling is again stratified by dataset so that every source is represented
in validation.
These validation examples are held fixed for the duration of the stage and
are not reused for training.

\paragraph{Early stopping.}
We monitor the validation loss once per fixed number of training steps and
terminate the current stage when the loss fails to improve for a specified
patience window.
If the best checkpoint occurs before the end of the stage, we roll back model
parameters to that checkpoint before proceeding to the next stage.
All random splits are generated with fixed seeds and logged to disk to enable
exact reproduction of data partitions.

\section{Training}

\subsection{Input Preprocessing \& Setup}
\label{app:setup}

During training, 
we rasterize each PDF page at 72\,DPI into a single image and feed it directly
to the Qwen2.5-VL NaViT-based vision encoder, preserving the original aspect
ratio and using the default maximum number of image tokens without additional
cropping, tiling, or high-resolution modes.

\paragraph{Backbone configuration.}
We fine-tune the public instruction-tuned checkpoints
(\texttt{Qwen2.5-VL-3B-Instruct} and \texttt{Qwen2.5-VL-7B-Instruct}) without
modifying the tokenizer or vocabulary.
All modules of the model are trainable: we do not freeze the vision encoder or
language backbone, and we do not introduce adapters or LoRA layers in the
GutenOCR models.

\paragraph{Context handling.}
We train and serve GutenOCR with long-context transcripts, using a
curriculum over sequence length described in Section~\ref{sec:training}.
Across stages, we cap the total sequence length (image prompt plus output
tokens) between roughly 2k and 16k tokens, always using a single page per
example with no dynamic packing across pages. When a target transcript would
exceed the stage-specific maximum, we truncate the output and discard the
remainder.

At inference time, we enforce the same total-context limits and cap the number
of generated tokens at 4{,}096 on external benchmarks and 16{,}384 in our
internal long-sequence evaluations.

\paragraph{Decoding and task routing.}
All reported results use deterministic greedy decoding (temperature $=0$,
no sampling) with the maximum new-token limits described above.
Task routing is purely prompt-based: the user specifies the desired task
(\texttt{reading}, \texttt{detection}, \texttt{localized\_reading}, or
\texttt{conditional\_detection}) and output schema
(\texttt{text}, \texttt{text2d}, \texttt{lines}, \texttt{paragraphs},
\texttt{BOX}, \emph{etc.}) in the prompt, and the same GutenOCR checkpoint
is used for all task families.
We do not employ any external head-selection logic or multi-model routing.

\subsection{Hyperparameters}
\label{app:hyperparams}

Table~\ref{tab:stage-hparams} summarizes the stage-specific optimization
settings used in all experiments.
Across stages, we fix the effective batch size to 128 samples
(16 per GPU device across 8$\times$H100) and use the same optimizer and
schedule:
AdamW~\cite{loshchilov2019decoupledweightdecayregularization} with weight
decay $0.01$ and maximum learning rate $1\times10^{-6}$, linearly warmed up
over the first $1{,}000$ updates and then linearly decayed to zero.
All modules are fully trainable and we use DeepSpeed ZeRO-3 for memory
sharding of parameters, optimizer states, and activations.
We train in bf16 with gradient checkpointing and clip all gradients to a global norm of 1.0.

\begin{table}[htb]
    \centering
    \small
    \begin{tabular}{c c c c c c p{4.5cm}}
    \toprule
    \textbf{Stage} &
    \textbf{Per-GPU} &
    \textbf{Grad} &
    \textbf{\#GPUs} &
    \textbf{Eff.\ batch} &
    \textbf{Length} &
    \textbf{Approx.\ dataset mixture} \\
    & \textbf{batch} & \textbf{accum} & & \textbf{size} & \textbf{range} & \\
    \midrule
    1  & 4 & 4  & 8 & 128 & $< 2$k
       & GL $\sim$48\%, IDL $\sim$45\%, SDG $\sim$6\%, T++ $\sim$0.4\% \\
    2  & 2 & 8  & 8 & 128 & $[2$k, $8$k$]$
       & IDL $\sim$95\%, T++ $\sim$5\% \\
    3a & 1 & 16 & 8 & 128 & $[2$k, $8$k$]$
       & PMD $\sim$32\%, IDL $\sim$65\%, T++ $\sim$3\% \\
    3b & 1 & 16 & 8 & 128 & $[8$k, $16$k$]$
       & PMD 100\% \\
    \bottomrule
    \end{tabular}
    \caption{
    Stage-specific optimization settings.
    Effective batch size is computed as (per-GPU batch) $\times$
    (gradient accumulation) $\times$ 8 GPUs.
    Dataset mixtures are approximate fractions of pages sampled per stage.
    SDG = SynthDoG Grounding, GL = Grounded \LaTeX{}, IDL = OCR-IDL,
    T++ = TabMe++, PMD = PubMed-OCR.
    }
    \label{tab:stage-hparams}
\end{table}

\subsection{System Prompt}
\label{app:system-prompt}

\begin{lstlisting}[style=prompt,basicstyle=\tiny]
Your task is to read and localize text data from documents and images.

GEOMETRY:
    - Coordinates: integer pixels; origin (0,0) top-left; [x1,y1,x2,y2] with x1<x2, y1<y2.
    - Clip all boxes to the image bounds; drop boxes with zero/negative area.
    - Reading order: read text in natural reading order: top-to-bottom, left-to-right.
    - Rotated/angled text: return the axis-aligned bounding box of the minimal enclosing rectangle (no rotated boxes).

TASK TYPES:
    - reading: a full-text reading task on the entire image.
    - localized_reading: read text within a specified bounding box in the image.
    - detection: detect text regions in the image without transcription.
    - conditional_detection: detect text regions in the image based on a provided text query.

OUTPUT TYPES:
    - TEXT: one plain string; collapse multiple spaces to one; preserve line breaks. Non-grounded output only.
    - TEXT2D: one plain string; preserve whitespace as layout cue (spaces + `\n` only; no coordinates). Non-grounded output only.
    - LINES: JSON array of objects, corresponding to line-by-line OCR: `{"text": string, "bbox": [x1,y1,x2,y2]}`. When locally reading, only return the text: `string'.
    - WORDS: JSON array of objects, corresponding to word-by-word OCR: `{"text": string, "bbox": [x1,y1,x2,y2]}`.
    - PARAGRAPHS: JSON array of objects, corresponding to paragraph-wise OCR: `{"text": string, "bbox": [x1,y1,x2,y2]}`. When locally reading, only return the text: `string`.
    - LATEX: JSON array of objects, corresponding to LaTeX expressions: `{"text": string, "bbox": [x1,y1,x2,y2]}`. When locally reading, only return the latex: `string`.
    - BOX: JSON array of bounding boxes only: `[ [x1,y1,x2,y2], ... ]`. For detection and conditional_detection tasks only.

OUTPUT FORMAT
    - For non-grounded outputs, return a string: 

        ```text
        Recognized text goes here.
        ```

        ```text2d
            ABSTRACT
        
        Recognition of text in a 2D layout.
        ```

        If the output is empty, return an empty string:
        
        ```text
        ```

        ```text2d
        ```

    - For grounded outputs, return a JSON array of objects when performing reading tasks.
      Each object is expected to have two keys: "text" and "bbox". 
      The "text" key is the what and the "bbox" key is the where.

        ```json
        [
            {"text": "First line of text", "bbox": [100, 200, 400, 250]},
            {"text": "Second line of text", "bbox": [100, 500, 400, 600]}
        ]
        ```

        ```json
        [
            {"text": "\\frac{a}{b}", "bbox": [525, 558, 755, 620]}
        ]
        ```

        If the output is empty, return an empty JSON array:

        ```json
        []
        ```
    - For detection tasks, return a JSON array of bounding boxes only.

        ```json
        [
            [100, 200, 400, 250],
            [100, 500, 400, 600]
        ]
        ```
    - For localized reading tasks, return the recognized text within the specified bounding box.

        ```text
        Recognized text within the bounding box.
        ```

        If no text is recognized within the bounding box, return an empty string:

        ```text
        ```
\end{lstlisting}

\section{Evaluation}
\label{app:evaluation}

\subsection{Parsing Structured Outputs}
\label{app:parsing}

Many baselines and all GutenOCR variants are prompted to emit JSON-structured
outputs (e.g., arrays of \texttt{\{"text", "bbox"\}} pairs).
In practice, LLM-generated JSON is frequently imperfect, so we employ a simple
repair-and-normalization pipeline before evaluation.

We first pass the raw model output to a tolerant JSON repair library
(\texttt{json\_repair}), which attempts to recover valid objects even in the
presence of unbalanced brackets, missing commas, comments, or extra text
around the JSON snippet.
If repair fails, the prediction is treated as invalid and assigned the maximal
error according to the conventions in Appendix~\ref{app:metrics}.

For successfully repaired outputs, we normalize key names and shapes:
keys whose names contain ``\texttt{text}'' or ``\texttt{label}'' (case
insensitive) are mapped to a canonical \texttt{text} field; keys containing
``\texttt{box}'' are mapped to \texttt{bbox}.
Lists of four numeric values are interpreted as bounding boxes and wrapped
into \texttt{\{"bbox": [x\_1,y\_1,x\_2,y\_2]\}} records when appropriate,
and any remaining coordinates are clipped to the image bounds.
This normalization allows us to treat structurally similar outputs from
different models in a uniform way while still penalizing irreparable or
semantically malformed predictions.

\subsection{Metrics}
\label{app:metrics}

Unless stated otherwise, all text metrics are computed on normalized strings
$\norm{\cdot}$.
The normalization operator applies Unicode NFKC, trims leading and trailing
whitespace, and collapses internal whitespace runs to a single space.
For \texttt{text2d} we preserve line breaks and intra-line spacing before
applying $\norm{\cdot}$ so that layout changes affect the score.

We summarize the formal definitions below for completeness.

\subsubsection{Text metrics}
\label{sec:metrics-text}

For a prediction $\hat{y}$ and reference $y$, let
$\lev(\cdot,\cdot)$ denote the standard Levenshtein edit distance and
$\len{\cdot}$ the character length.
The character error rate (CER) is
\[
\mathrm{CER}(\hat{y},y)=
\frac{\lev\bigl(\norm{\hat{y}},\norm{y}\bigr)}%
     {\max\!\bigl(1,\ \max\{\len{\norm{\hat{y}}},\,\len{\norm{y}}\}\bigr)}.
\]
Normalizing by the length of the longer string keeps
$\mathrm{CER}\in[0,1]$ even when one side is empty or much shorter than the
other, which is useful under noisy OCR-derived labels.

For word error rate (WER), we tokenize $\norm{y}$ into words using
whitespace and let $N=\words{\norm{y}}$ be the number of reference words.
Let $S,D,I$ be the number of substitutions, deletions, and insertions,
respectively, when aligning $\norm{\hat{y}}$ to $\norm{y}$ at the word level.
Then
\[
\mathrm{WER}(\hat{y},y)=\frac{S+D+I}{\max(1,N)}.
\]

\subsubsection{Detection metrics}
\label{sec:metrics-detection}

Let $P=\{b_i\}_{i=1}^m$ be the set of predicted boxes and
$G=\{g_j\}_{j=1}^n$ the ground-truth boxes, with each box
$b=[x_1,y_1,x_2,y_2]$ in image pixel coordinates.
We define the intersection-over-union
\[
\IoU(b,g)=\frac{\area(b\cap g)}{\area(b\cup g)}.
\]

For a threshold $\tau$ (we use $\tau=0.5$ throughout), we construct a
one-to-one matching $M_\tau\subseteq P\times G$ using only pairs with
$\IoU\ge\tau$ (Hungarian matching with IoU as affinity).
We then define
\[
\tp_\tau=\lvert M_\tau\rvert,\quad
\fp_\tau=m-\tp_\tau,\quad
\fn_\tau=n-\tp_\tau.
\]

Precision, recall, and F1 at threshold $\tau$ are
\[
\text{Prec}_\tau=
\frac{\tp_\tau}{\max(1,\tp_\tau+\fp_\tau)},\quad
\text{Rec}_\tau=
\begin{cases}
1, & |G|=0,\\[0.25em]
\tp_\tau/|G|, & |G|>0,
\end{cases}
\]
\[
\text{F1}_\tau=
\frac{2\,\text{Prec}_\tau\,\text{Rec}_\tau}{\max(1,\text{Prec}_\tau+\text{Rec}_\tau)}.
\]

In the main text we report $\mathrm{F1@0.5}\equiv\text{F1}_{\tau=0.5}$ and
$\mathrm{Recall@0.5}\equiv\text{Rec}_{\tau=0.5}$.

\subsubsection{End-to-end metrics (detection + recognition)}
\label{sec:metrics-e2e}

For structured reading tasks that produce box--text pairs
$P=\{(b_i,\hat{t}_i)\}$ and $G=\{(g_j,t_j)\}$ we use two end-to-end
metrics that reuse the detection matching $M_{0.5}$ from
Section~\ref{sec:metrics-detection}.

\paragraph{mCER@0.5 (recognition given correct localization).}
Let $M_{0.5}$ be the one-to-one matching between predicted and ground-truth
boxes at IoU~$\ge 0.5$.
For each $(i,j)\in M_{0.5}$ we compute a per-span CER
\[
\mathrm{cer}_{i,j}=
\frac{\lev\bigl(\norm{\hat{t}_i},\norm{t_j}\bigr)}%
     {\max\!\bigl(1,\ \max\{\len{\norm{\hat{t}_i}},\,\len{\norm{t_j}}\}\bigr)}.
\]
mCER@0.5 is the mean of $\mathrm{cer}_{i,j}$ over $(i,j)\in M_{0.5}$, with
edge cases handled as described below.
Because unmatched ground-truth boxes do not contribute directly to this
average, mCER@0.5 isolates recognition quality conditioned on successful
localization; coverage is accounted for by the detection recall.

\paragraph{$\text{CER}_{\text{e2e}}$ (page-level fidelity).}
To assess the joint effect of localization, reading order, insertions, and
deletions, we form page-level strings from $P$ and $G$ as follows.
We sort boxes in a deterministic reading order (by the $y_1$ coordinate of
their top edge, then by $x_1$ left-to-right), concatenate the corresponding
texts with line breaks and layout-sensitive whitespace (as in the
\texttt{text2d} construction), and finally apply $\norm{\cdot}$ to both
page strings.
We then compute CER between these two concatenated strings.
$\text{CER}_{\text{e2e}}$ is therefore sensitive to missing or spurious
lines, scrambled ordering, and layout-preserving whitespace errors in
addition to pure recognition mistakes.

\subsubsection{Edge cases and invalid predictions}

We adopt the following conventions to keep metrics well defined:

\begin{itemize}
  \item \textbf{Empty text.}
        If both normalized strings are empty, we set $\mathrm{CER}=0$ and
        $\mathrm{WER}=0$.
        If exactly one side is empty, we set $\mathrm{CER}=1$ and
        $\mathrm{WER}=1$.

  \item \textbf{Detection with no boxes.}
        \begin{itemize}
          \item $P=\varnothing$ and $G=\varnothing$:
                $(\text{Prec},\text{Rec},\text{F1})=(1,1,1)$.
          \item $G=\varnothing$ and $P\neq\varnothing$:
                $\text{Rec}=1$, $\text{Prec}=0$, so $\text{F1}=0$.
          \item $P=\varnothing$ and $G\neq\varnothing$:
                $\text{Rec}=0$ and $\text{F1}=0$.
        \end{itemize}

  \item \textbf{mCER@0.5 with no matches.}
        If $|M_{0.5}|>0$, mCER@0.5 is the average over matched pairs.
        If $P=\varnothing$ and $G=\varnothing$, we set $\mathrm{mCER@0.5}=0$.
        In all other cases with $|M_{0.5}|=0$ (e.g., detector fires but never
        overlaps ground truth, or misses all ground-truth boxes) we set
        $\mathrm{mCER@0.5}=1$.

  \item \textbf{Invalid structured predictions.}
        If a model output cannot be repaired into valid JSON or yields no
        usable boxes/text after normalization (Appendix~\ref{app:parsing}),
        we treat it as a maximal-error prediction:
        $\mathrm{CER}=1$, $\mathrm{WER}=1$.
        For structured tasks we additionally set
        $\mathrm{mCER@0.5}=1$ and $\text{CER}_{\text{e2e}}=1$, and assign
        $\mathrm{F1@0.5}=0$ and $\mathrm{Recall@0.5}=0$ for the associated
        detection metrics.
\end{itemize}

\paragraph{Composite grounded OCR score.}
For the composite scores reported in Section~\ref{sec:in-domain}, we map
each reading error $\epsilon\in[0,1]$ (CER, WER, mCER@0.5, or
$\text{CER}_{\text{e2e}}$) to a per-task score $1-\epsilon$, and average
these scores with the corresponding detection and conditional-detection
F1@0.5 values over the task families in Table~\ref{tab:metrics-cheatsheet}.
The resulting composite lies in $[0,1]$ and weights reading and detection
tasks equally.

\section{Additional Benchmark Details}
\label{app:benchmarks}

\subsection{Fox Benchmark Details}
\label{app:fox}

This section supplements the main Fox results in Section~\ref{sec:fox} and
Table~\ref{tab:fox-bench} with task configuration, prompting details, and a
more detailed discussion of the Page F1 vs.\ Page CER trade-off and of the
region, line, and color-guided subtasks.

\paragraph{Tasks and filtering.}
Fox~\cite{liu2024fox} defines a collection of multi-page document tasks that
mix OCR, retrieval, and visual reasoning. We restrict our evaluation to the
four English OCR-style subtasks that align with GutenOCR's training objectives:
(i) page OCR, (ii) region-level OCR, (iii) line-level OCR, and (iv)
color-guided OCR. We follow the official Fox splits and evaluation scripts,
but filter pages to those whose language metadata is purely English.

\paragraph{Output canonicalization.}
For the page OCR subtask, we score predictions as plain text. In practice, some
models emit lightweight structure or markup (e.g., JSON wrappers, Markdown, or
HTML) rather than a raw transcript. To make evaluation robust to these surface
forms, we apply a uniform post-processing step that attempts to detect and
render such outputs into a single text string prior to scoring (Markdown is
rendered via \texttt{markdown2}, HTML is stripped via \texttt{BeautifulSoup},
and JSON outputs are parsed and linearized into a text representation).
This canonicalization is applied identically to all models, but it may mildly
benefit systems that produce well-formed structured/marked-up outputs by
allowing their textual content to be recovered instead of being penalized for
formatting alone.

\paragraph{Prompting and decoding.}
All models, including GutenOCR and Qwen-based baselines, are evaluated under a
unified prompting scheme. For each Fox subtask, we use the official benchmark
prompts and output-format instructions as provided by the evaluation scripts.
For the region- and line-level OCR tasks (which include an explicit pointer
region), the only model-specific adaptation we apply is a deterministic
conversion of the pointer box into the coordinate convention expected by the
model (e.g., absolute pixel coordinates vs.\ normalized coordinates). Beyond
this box-format normalization, prompts are otherwise unchanged across models.

For baselines, we apply a minimal OCR-focused system prompt (Listing~\ref{lst:fox_system_prompt})
to encourage strict format compliance (e.g., emitting raw JSON when requested).
For GutenOCR, we preserve the exact system prompt used during training (see
Appendix~\ref{app:system-prompt}), without modification.

All results are produced with deterministic greedy decoding (temperature $=0$,
no sampling) and a fixed maximum new-token budget.
\begin{lstlisting}[style=prompt,basicstyle=\tiny,label={lst:fox_system_prompt}]
You are an OCR assistant. Read and transcribe text from the provided document image.

Follow the user's instructions exactly.
Do not add explanations or extra text beyond the requested output.
\end{lstlisting}

\paragraph{Page metrics: Page F1 vs.\ Page CER.}
Fox reports an order-agnostic \emph{Page F1} metric that measures the overlap
between predicted and reference tokens after text normalization and an
order-sensitive edit-distance metric. In our tables, we retain Page F1 and
replace the latter with our normalized character error rate (CER;
Section~\ref{sec:metrics}). Intuitively:

\begin{itemize}
    \item \textbf{Page F1} is forgiving of reading order and extra whitespace.
    It treats the page as a bag (or multiset) of tokens and measures how many
    reference tokens appear somewhere in the prediction, and vice versa.
    \item \textbf{Page CER} is stricter: predictions are compared to the
    canonical Fox linearization as a single string, so reordering blocks,
    inserting extra text, or missing spans all increase the error.
\end{itemize}

GutenOCR is trained to emit layout-sensitive \texttt{text2d} and structured
line/paragraph JSON outputs, not a single canonical page-to-Markdown format.
On Fox pages, it therefore tends to obey 2D layout cues (columns, tables,
headings, whitespace) rather than the specific reading order chosen by the
benchmark. As a result, GutenOCR-3B and GutenOCR-7B achieve very high Page F1
(close to or above their Qwen2.5-VL backbones) while exhibiting substantially
worse Page CER: they read essentially the right content, but prefer to do so in a layout-driven
order that diverges from Fox's reference markup.

In contrast, OCR-specialized baselines like Nanonets-OCR2-3B and
olmOCR~2 are explicitly optimized for page-to-Markdown fidelity and thus
better match Fox's canonical order, yielding lower Page CER at similar or
slightly lower Page F1. We view this discrepancy as reflecting a difference
in target behavior rather than a pure weakness: GutenOCR prioritizes
layout-preserving, grounded front-end outputs, whereas Fox's page metric
rewards adherence to a particular markup convention.

\paragraph{Region- and line-level OCR.}
Region and line subtasks probe the ability to ``focus anywhere'' on the
document. Fox provides, for each query, either a region pointer (a box
highlighted on the page) or a line pointer, together with a target transcript.
We evaluate all models by cropping the indicated region/line from the page,
passing the full page plus pointer to the model, and computing CER between the
normalized prediction and reference text.

GutenOCR is closest to these tasks in spirit: it is explicitly trained for
region-aware reading and localized reading given a box. As reported in
Table~\ref{tab:fox-bench}, this specialization yields large gains over the
Qwen2.5-VL backbones on both region and line tasks, and GutenOCR-3B attains
the best region-level CER among all evaluated models, slightly outperforming
the dedicated Fox model itself. The line task is somewhat extrapolative for
GutenOCR (we do not train on explicit line-pointer prompts), but we still see
substantial improvements, which we attribute to transfer from region-level
detection and localized reading: once the model can reliably find and read
arbitrary subregions, line pointers become a natural special case.

\paragraph{Color-guided OCR and catastrophic forgetting.}
The color-guided task asks the model to read only text in a colored box
(red/green/blue) while ignoring all other text.
Fox and several OCR-specialized baselines perform strongly here; Qwen2.5-VL-7B
also demonstrates a reasonable zero-shot ability to interpret color-coded
pointers, likely inherited from generic vision-and-grounding pretraining.

GutenOCR, by contrast, performs poorly on this subtask (high CER for both 3B
and 7B). We do not include any color-guided supervision in our training
mixture, and the OCR-centric fine-tuning appears to overwrite the heuristics
the base VLMs had learned for color-based selection.
Qualitatively, we frequently observe GutenOCR interpreting prompts literally
(e.g., trying to read the words ``red box'' from the page) or returning
structured box coordinates instead of the requested text.
This is a classic example of narrow catastrophic forgetting: the
documentation-focused curriculum strengthens grounded OCR in the regime we
care about (full-page, region, and localized reading) while eroding an
orthogonal emergent behavior (color-conditioned region selection) that was
never exercised during fine-tuning.

\paragraph{Qualitative examples.}
Figure~\ref{fig:fox-model-comparison} illustrates typical Fox behaviors for
Qwen2.5-VL-3B vs.\ GutenOCR-3B:

\begin{itemize}
    \item \textbf{Region/line success.}
    In panels (a) and (b), Qwen2.5-VL-3B either misses or badly misreads the
    targeted snippet, while GutenOCR returns the correct localized text with
    near-zero CER, showing transfer of its in-domain grounding abilities.
    \item \textbf{Color failure.}
    Panel (c) shows a case where Qwen2.5-VL-3B perfectly follows the color
    box, whereas GutenOCR essentially ignores the color cue.
    \item \textbf{Content vs.\ structure.}
    Panel (d) provides a concrete Page F1 vs.\ Page CER example: GutenOCR
    captures almost all of the reference tokens (high F1) but streams them in
    a layout-driven order that diverges from the Fox linearization, leading to
    a higher CER than the backbone despite better underlying reading.
\end{itemize}

These examples match the quantitative pattern in
Table~\ref{tab:fox-bench}: GutenOCR behaves like a grounded OCR
front-end---strong on region and line reading, layout-sensitive in its page
outputs, and relatively agnostic to color cues---rather than like a
page-to-Markdown system tuned for a single canonical reading order.

\subsection{OmniDocBench v1.5}
\label{app:omnidoc}

OmniDocBench v1.5~\cite{ouyang2025omnidocbenchbenchmarkingdiversepdf} is a
multi-domain document benchmark built from PDF pages with dense annotations
for layout regions, text spans, formulas, and tables.
The corpus covers nine page types---books, slides, financial reports,
textbooks, exam papers, magazines, academic papers, handwritten notes, and
newspapers---with a v1.0$\rightarrow$1.5 revision that expands coverage and
refines markup conventions.\footnote{See the v1.0$\rightarrow$1.5 changelog in
the OmniDocBench supplement.}
The official tasks focus on end-to-end markup generation (e.g., HTML/Markdown)
and report a combination of edit-distance and structure-aware metrics
for text, formulas, and tables.

Our production setting targets grounded text+box outputs rather than canonical
markup, so we use OmniDocBench as a testbed for three task-specific behaviors:
\emph{text recognition}, \emph{formula recognition}, and \emph{text detection}.
Throughout, we restrict to pages whose language tag is purely English.

\paragraph{Text recognition subset.}
For text recognition we use the OmniDocBench OCR subset, which provides cropped
text spans together with attributes such as language, background type, and
rotation.
Each span is paired with a reference transcript, and evaluation is based on
component-level edit distance.
We follow the official setup and compute character error rate (CER) on either
(i) full-page concatenations (page-level OCR) or (ii) individual cropped spans
(component-level OCR), grouping results by attributes such as background
(white, single-color, multi-color) and page type.
In all cases we apply the same string normalization $\norm{\cdot}$ as in
Appendix~\ref{app:metrics} before computing CER.

\paragraph{Formula recognition subset.}
OmniDocBench includes a dedicated formula subset with \LaTeX{} annotations for
rendered equations.
The official benchmark evaluates formulas using the Character Detection
Matching (CDM) metric~\cite{wang2025imagetexttransformingformula}, which
compares rendered strings at the symbol level, together with normalized
edit distance.
We adopt this setup for the English subset:
for each cropped formula we render both the reference and predicted \LaTeX{}
and compute (i) CER between normalized strings, and (ii) CDM as a
render-aware similarity score (higher is better).
Results in the main text (Table~\ref{tab:omnidoc-formula-recognition}) break
these metrics down by page type to surface domain-specific trends.

\paragraph{Layout and text-detection subset.}
For layout, OmniDocBench provides text spans defined at roughly line level,
together with higher-level layout regions such as tables and formula blocks.
Crucially for our purposes, the text spans form a \emph{partial} annotation of
visible text:
\begin{itemize}
    \item Many table cells are represented only through a table object rather
          than individual text spans.
    \item Certain formulas and labels are marked with special tags
          (e.g., \texttt{Equation Ignore}) to exclude them from text-based
          evaluation.
    \item Decorative or low-salience text may be omitted entirely.
\end{itemize}
GutenOCR, by contrast, is trained under a ``detect all readable text''
ontology: it aims to place line-level boxes on any legible text, including
table cells, equation-like expressions, and decorative labels.
As a result, a one-to-one mapping between OmniDocBench spans and our notion of
``text line'' does not exist.

\paragraph{Recall-only detection protocol.}
Because OmniDocBench text spans cover only a subset of the text that GutenOCR
legitimately detects, we cannot reliably treat non-overlapping predictions as
false positives.
Any predicted box with no matching annotated span might be:
(i) a true positive under our ontology (e.g., content inside a table cell), or
(ii) a genuine hallucination; the benchmark does not distinguish between these
cases.
This makes precision and F1 ill-defined for our setting.

To avoid over-interpreting partial labels, we treat OmniDocBench text spans as
a \emph{trusted but incomplete} ground-truth set and evaluate only recall at
IoU~$\geq 0.5$:
\begin{quote}
When OmniDocBench says ``there is text here'', how often does our detector
cover it?
\end{quote}
Concretely, we compute R@0.5 by matching predicted and ground-truth boxes
using IoU~$\geq 0.5$ (as in Appendix~\ref{app:metrics}) and report the fraction
of annotated spans that receive a match.
We do not report precision or F1 on OmniDocBench text detection.

Qualitatively, this protocol corresponds to counting red (annotated) boxes
that are covered by blue (predicted) boxes in examples such as
Figure~\ref{fig:omnidoc-text-det-example}, while ignoring extra blue boxes on
unannotated but readable text (table cells, decorative labels, etc.).
Higher recall under this metric therefore indicates better coverage of the
annotated subset, without making claims about behavior on the unannotated
portion of the page.

\section{Models}
\label{app:models}

We group baselines into two categories:
(i) general-purpose Qwen-based VLMs used in a prompting-only setting, and
(ii) OCR-specialized or layout-focused models built on Qwen or other
backbones.
A high-level overview of all models, including parameter counts, backbones, and
whether their public interfaces expose coordinates, is given in
Table~\ref{tab:model-summary}.

\paragraph{Qwen2.5-VL}
Qwen2.5-VL~\cite{bai2025qwen25vltechnicalreport} is a general-purpose
vision--language model trained on a mixture of image--text,
document-understanding, OCR, and spatial-grounding tasks, including referring
expressions and box/point localization.
In its native interface, spatial outputs are expressed as absolute pixel
coordinates in the input image frame (e.g., bounding boxes
\(([x_1,y_1,x_2,y_2])\)), which matches our detection and localized-reading
conventions.
We use the off-the-shelf instruct models (3B/7B) as prompting-only baselines
for our reading, detection, conditional-detection, and localized-reading tasks,
specifying the desired output schema (plain text, layout-sensitive text, or
JSON with boxes) in the prompt.

\paragraph{Qwen3-VL}
Qwen3-VL-8B~\cite{bai2025qwen3vltechnicalreport} is an 8B-parameter
vision--language model trained on large-scale OCR, document parsing,
long-document VQA, and 2D/3D grounding and counting data, in addition to
generic image--text and VQA corpora.
Spatial supervision is provided through both bounding-box and point
annotations, represented in a normalized coordinate system scaled to
$[0,1000]$ along each image axis, which we map to pixel coordinates for
evaluation.
We use the off-the-shelf Qwen3-VL-8B instruct model as another prompting-only
baseline, again prompting for the desired output schema (plain text,
layout-aware text, or JSON with boxes).

\paragraph{Nanonets OCR2-3B}
Nanonets-OCR2-3B~\cite{Nanonets-OCR2} is a 3B-parameter OCR-specialized
vision--language model obtained by fine-tuning Qwen2.5-VL-3B-Instruct on a
$>\!3$M-page corpus of synthetic and manually annotated documents for
image-to-Markdown OCR and document-focused visual question answering.
The model takes a document page as input and produces a single structured
Markdown string with semantic tags for equations, tables, images, signatures,
watermarks, checkboxes, page numbers, and flowcharts, but does not expose
bounding-box coordinates or region-level grounding outputs.
In our evaluation, we primarily use Nanonets-OCR2-3B as a strong baseline for
page-level reading and tag-level content extraction, and additionally probe its
performance on our coordinate-based detection and grounding tasks to assess how
much of the underlying Qwen2.5-VL-3B spatial reasoning ability is retained
after OCR specialization.

\paragraph{olmOCR 2}
olmOCR 2~\cite{poznanski2025olmocr2unittest} is an OCR-specialized 7B
vision--language model obtained by further fine-tuning
Qwen2.5-VL-7B-Instruct with supervised training and reinforcement learning
with verifiable rewards (RLVR) on synthetic HTML-rendered pages, using unit
tests that score only the correctness of a linearized textual transcript
(reading order, tables, math rendering, etc.).
The public system takes a rasterized PDF page as input and emits YAML whose
body is plain Markdown text, without explicit bounding boxes or coordinates.
We therefore regard olmOCR 2 as a strong baseline for page-level reading and
use our evaluation to quantify how much of the base model’s detection and
referring ability is lost under OCR-focused specialization.

\paragraph{Infinity-Parser-7B}
Infinity-Parser-7B~\cite{wang2025infinityparserlayoutaware} is a layout-aware
document-parsing model obtained by fine-tuning Qwen2.5-VL-7B on the
Infinity-Doc-55K corpus of scanned pages paired with structured Markdown and
table/formula annotations, using a reinforcement-learning objective that
combines normalized edit distance, paragraph-count accuracy, and reading-order
preservation.
The public model takes page images as input and produces Markdown (with region
segments wrapped in \texttt{<ele>} tags) and HTML tables, but does not emit
bounding boxes or coordinates.
We treat Infinity-Parser-7B as a strong baseline for page-level reading and
layout-sensitive parsing, and additionally evaluate it on our grounding tasks
to measure how much of the Qwen2.5-VL-7B base model’s detection and referring
ability remains after layout-focused RL tuning.

\paragraph{Chandra OCR}
Chandra~\cite{Paruchuri_2025} is a layout-preserving OCR system built on top
of the Qwen3-VL vision--language model, trained to convert full document pages
into structured HTML/Markdown/JSON while handling complex layouts,
handwriting, tables, forms, math, and images.
The model outputs HTML comprising block-level \texttt{<div>} elements
annotated with semantic labels (e.g., text, image, figure, table) and
bounding boxes in a normalized $1024\times1024$ coordinate system via
\texttt{data-bbox} attributes, which the library rescales to pixel coordinates
for downstream processing.
We primarily use Chandra as a baseline for page-level reading and block-level
detection in our taxonomy, and additionally test it on our finer-grained
line-level and conditional grounding tasks, which go beyond the capabilities
explicitly targeted by its training.

\paragraph{DeepSeek-OCR}
DeepSeek-OCR~\cite{wei2025deepseekocrcontextsopticalcompression} is a
3B-parameter vision--language OCR model composed of a custom high-resolution
vision encoder (DeepEncoder) and a DeepSeek-3B-MoE decoder, trained on a large
mixture of document OCR, synthetic ``OCR 2.0'' (charts, chemical formulas,
geometry), natural-scene OCR, general vision (caption/detection/grounding),
and text-only data.
Document pages are supervised with both coarse text-only labels and fine layout
annotations in which each paragraph is preceded by its region label and
bounding-box coordinates normalized to 1{,}000 bins, enabling the model to
emit layout-aware text paired with coordinates as well as plain transcripts.
We use DeepSeek-OCR as a baseline for page-level reading and layout-sensitive
parsing, and further evaluate it on our detection and conditional grounding
tasks to probe its referring and object-localization capabilities.

\paragraph{PaddleOCR-VL}
PaddleOCR-VL-0.9B~\cite{cui2025paddleocrvlboostingmultilingualdocument} is a
compact, document-specialized VLM trained for element-level recognition (OCR,
tables, formulas, charts) but not for layout detection or box prediction.
In the full PaddleOCR system, a separate layout model (PP-DocLayoutV2) first
predicts element types, boxes, and reading order; cropped regions are then fed
to PaddleOCR-VL-0.9B, which is trained---via a 29M pretraining set plus 2.7M
instruction samples---to output structured text such as Markdown/JSON, OTSL
tables, and \LaTeX{} formulas.
In our evaluation we explicitly use only the PaddleOCR-VL-0.9B VLM as another
OCR/document-parsing baseline, and do not rely on their separate
layout-analysis stage.

\section{Additional Results}

\subsection{Detailed In-Domain Evaluation}
\label{app:detailed-evaluation}

\subsubsection{Reading}
\label{app:detailed-evaluation-reading}

\paragraph{Structured, line-based reading.}

\begin{table}[htb]
    \centering\small
    \begin{tabular}{lc cc cc cc cc}
    \toprule
    \multirow{2}{*}{\textbf{Model}} &
    \multirow{2}{*}{\textbf{Stage}} &
    \multicolumn{2}{c}{\textbf{OCR-IDL}} &
    \multicolumn{2}{c}{\textbf{TabMe++}} &
    \multicolumn{2}{c}{\textbf{PubMed-OCR}} &
    \multicolumn{2}{c}{\textbf{AVG}} \\
    & & CER$_{\text{e2e}}$ & mCER & CER$_{\text{e2e}}$ & mCER & CER$_{\text{e2e}}$ & mCER & CER$_{\text{e2e}}$ & mCER \\
    \midrule
    Qwen2.5-VL-3B  &    & 0.372 & 0.418 & 0.284 & 0.247 & 0.479 & 0.559 & 0.379 & 0.408 \\
    \quad GutenOCR-3B & 1  & \underline{0.140} & 0.068 & 0.175 & 0.054 & 0.235 & 0.113 & \underline{0.183} & 0.078 \\
    \quad GutenOCR-3B & 2  & 0.144 & \underline{0.065} & 0.172 & \underline{0.048} & 0.238 & 0.116 & 0.185 & 0.076 \\
    \quad GutenOCR-3B & 3a & 0.153 & 0.071 & \underline{0.170} & 0.051 & \underline{0.233} & \underline{0.081} & 0.185 & \underline{0.067} \\
    \quad GutenOCR-3B & 3b & 0.191 & 0.093 & 0.197 & 0.072 & 0.235 & 0.077 & 0.208 & 0.081 \\
    \cmidrule{3-10}
    Qwen2.5-VL-7B  &    & 0.638 & 0.196 & 0.651 & 0.129 & 0.611 & 0.379 & 0.633 & 0.235 \\
    \quad GutenOCR-7B & 1  &
        \underline{\textbf{0.114}} &
        \underline{\textbf{0.058}} &
        \underline{\textbf{0.149}} &
        \underline{\textbf{0.045}} &
        0.159 & 0.099 &
        \underline{\textbf{0.141}} & 0.067 \\
    \quad GutenOCR-7B & 2  & 0.128 & 0.060 & 0.156 & 0.047 & 0.177 & 0.107 & 0.154 & 0.071 \\
    \quad GutenOCR-7B & 3a & 0.128 & 0.062 & 0.159 & 0.047 & 0.155 & 0.070 & 0.147 & \underline{\textbf{0.060}} \\
    \quad GutenOCR-7B & 3b & 0.169 & 0.089 & 0.188 & 0.069 & \underline{\textbf{0.150}} & \underline{\textbf{0.060}} & 0.169 & 0.073 \\
    \bottomrule
    \end{tabular}
    \caption{
        Structured, line-based full-page reading on OCR-IDL, TabMe++, and PubMed-OCR.
        CER$_\text{e2e}$ is the page-level character error rate on the concatenated transcript; mCER is computed only on lines whose boxes are correctly matched.
        All values are errors (lower is better, $\downarrow$).
        Within each column, bold numbers indicate the best value across all models; underlined numbers are best within each backbone cohort (3B vs.\ 7B).
        Entries that are both bold and underlined are best both overall and within their cohort.
    }
    \label{tab:reading-lines}
\end{table}

GutenOCR converts both Qwen2.5-VL backbones from “good reader, bad paginator’’ into stable structured readers.
On the 3B backbone, Stage~1 roughly halves average CER$_\text{e2e}$ and cuts mCER by about 80\% relative to Qwen2.5-VL-3B, with Stages~2 and~3a mostly refining per-line recognition (mCER) at similar page-level error.
On the 7B backbone, Stage~1 achieves the lowest overall CER$_\text{e2e}$ and Stage~3a the lowest mCER, eliminating the large gap between strong per-line accuracy and poor page assembly seen in the raw Qwen2.5-VL-7B model.
Stage~3b over-specializes to PubMed-OCR, improving that domain at a modest cost to the business datasets.

\begin{table}[htb]
    \centering\small
    \begin{tabular}{lc cc cc cc cc}
    \toprule
    \multirow{2}{*}{\textbf{Model}} &
    \multirow{2}{*}{\textbf{Stage}} &
    \multicolumn{2}{c}{\textbf{OCR-IDL}} &
    \multicolumn{2}{c}{\textbf{TabMe++}} &
    \multicolumn{2}{c}{\textbf{PubMed-OCR}} &
    \multicolumn{2}{c}{\textbf{AVG}} \\
    &  & CER & WER & CER & WER & CER & WER & CER & WER \\
    \midrule
    Qwen2.5-VL-3B  &    & 0.508 & 0.535 & 0.438 & 0.503 & 0.577 & 0.735 & 0.508 & 0.591 \\
    \quad GutenOCR-3B & 1  &
        \underline{0.186} & \underline{0.263} &
        \underline{0.207} & \underline{0.284} &
        0.259 & 0.528 &
        \underline{0.218} & 0.358 \\
    \quad GutenOCR-3B & 2  & 0.201 & 0.274 & 0.212 & 0.287 & \underline{0.257} & 0.527 & 0.223 & 0.363 \\
    \quad GutenOCR-3B & 3a & 0.202 & 0.277 & 0.209 & 0.286 & 0.260 & \underline{0.346} & 0.224 & \underline{\textbf{0.303}} \\
    \quad GutenOCR-3B & 3b & 0.230 & 0.446 & 0.234 & 0.438 & 0.296 & 0.359 & 0.253 & 0.414 \\
    \cmidrule{3-10}
    Qwen2.5-VL-7B  &    & 0.320 & 0.353 & 0.262 & 0.340 & 0.418 & 0.629 & 0.333 & 0.441 \\
    \quad GutenOCR-7B & 1  & 0.181 & \underline{\textbf{0.244}} & \underline{\textbf{0.198}} & \underline{\textbf{0.275}} & 0.228 & 0.505 & \underline{\textbf{0.198}} & 0.341 \\
    \quad GutenOCR-7B & 2  & 0.181 & 0.256 & 0.209 & 0.285 & 0.240 & 0.514 & 0.210 & 0.352 \\
    \quad GutenOCR-7B & 3a &
        \underline{\textbf{0.178}} & 0.253 &
        0.206 & 0.282 &
        \underline{\textbf{0.222}} & 0.343 &
        0.202 & \underline{0.297} \\
    \quad GutenOCR-7B & 3b & 0.221 & 0.448 & 0.247 & 0.459 & 0.254 & \underline{\textbf{0.312}} & 0.241 & 0.406 \\
    \bottomrule
    \end{tabular}
    \caption{
        Full-page reading with linearized text output.
        We report character error rate (CER) and word error rate (WER), lower is better ($\downarrow$).
        Bold numbers indicate the best value in each column across all models; underlined numbers are best within each backbone cohort.
    }
    \label{tab:reading-text}
\end{table}

Table~\ref{tab:reading-text} shows plain-text OCR where each page is reduced to a single transcript.
Scaling from 3B to 7B already helps substantially, but GutenOCR fine-tuning dominates: all GutenOCR variants reduce CER and WER versus their backbones.
Stage~1 gives the lowest average CER for both sizes, while Stage~3a trades a small amount of CER for noticeably better WER, especially on PubMed-OCR.
Stage~3b again behaves as a PubMed-OCR specialist.
Overall, GutenOCR-7B Stage~3a is the strongest linear-text reader, with GutenOCR-3B Stage~1/3a as competitive, cheaper options.

\paragraph{Layout-sensitive \texttt{text2d}.}

\begin{table}[htb]
    \centering\small
    \begin{tabular}{lc cc cc cc cc}
    \toprule
    \multirow{2}{*}{\textbf{Model}} &
    \multirow{2}{*}{\textbf{Stage}} &
    \multicolumn{2}{c}{\textbf{OCR-IDL}} &
    \multicolumn{2}{c}{\textbf{TabMe++}} &
    \multicolumn{2}{c}{\textbf{PubMed-OCR}} &
    \multicolumn{2}{c}{\textbf{AVG}} \\
    &  & CER & WER & CER & WER & CER & WER & CER & WER \\
    \midrule
    Qwen2.5-VL-3B &     & 0.623 & 0.545 & 0.505 & 0.397 & 0.623 & 0.710 & 0.584 & 0.562 \\
    \quad GutenOCR-3B & 1  &
        \underline{0.234} & \underline{0.249} &
        \underline{0.287} & \underline{0.272} &
        0.305 & 0.520 &
        \underline{0.276} & 0.347 \\
    \quad GutenOCR-3B & 2  & 0.254 & 0.264 & 0.290 & 0.279 & \underline{0.301} & 0.517 & 0.282 & 0.353 \\
    \quad GutenOCR-3B & 3a & 0.258 & 0.267 & 0.293 & 0.284 & 0.327 & \underline{0.350} & 0.293 & \underline{\textbf{0.300}} \\
    \quad GutenOCR-3B & 3b & 0.318 & 0.451 & 0.346 & 0.451 & 0.370 & 0.372 & 0.345 & 0.424 \\
    \cmidrule{3-10}
    Qwen2.5-VL-7B &     & 0.514 & 0.363 & 0.473 & 0.343 & 0.580 & 0.670 & 0.522 & 0.459 \\
    \quad GutenOCR-7B & 1  &
        \underline{\textbf{0.213}} & \underline{\textbf{0.229}} &
        \underline{\textbf{0.274}} & \underline{\textbf{0.260}} &
        \underline{\textbf{0.276}} & 0.503 &
        \underline{\textbf{0.255}} & 0.331 \\
    \quad GutenOCR-7B & 2  & 0.223 & 0.245 & 0.282 & 0.277 & 0.284 & 0.510 & 0.263 & 0.344 \\
    \quad GutenOCR-7B & 3a & 0.232 & 0.251 & 0.291 & 0.281 & 0.318 & 0.377 & 0.280 & \underline{0.303} \\
    \quad GutenOCR-7B & 3b & 0.290 & 0.458 & 0.344 & 0.473 & 0.336 & \underline{\textbf{0.327}} & 0.323 & 0.419 \\
    \bottomrule
    \end{tabular}
    \caption{
        Full-page reading with layout-preserving \texttt{text2d} output.
        We report CER and WER, lower is better ($\downarrow$).
        Bold numbers denote the best value in each column; underlined numbers are best within each backbone cohort.
    }
    \label{tab:reading-text2d}
\end{table}

Table~\ref{tab:reading-text2d} evaluates layout-preserving \texttt{text2d}, which is stricter than linear text because whitespace carries layout information.
Both GutenOCR backbones roughly halve CER and WER relative to their Qwen baselines.
Stage~1 gives the lowest CER on average, while Stage~3a improves WER by better handling of spaces and line breaks, again especially on PubMed-OCR.
After fine-tuning, differences between GutenOCR-3B and GutenOCR-7B in this setting are small compared to the gains over the original VLMs.

\subsubsection{Localized Reading}
\label{app:detailed-evaluation-local}

\begin{table}[htb]
    \centering\small
    \begin{tabular}{lc cc cc cc cc}
    \toprule
    \multirow{2}{*}{\textbf{Model}} &
    \multirow{2}{*}{\textbf{Stage}} &
    \multicolumn{2}{c}{\textbf{OCR-IDL}} &
    \multicolumn{2}{c}{\textbf{TabMe++}} &
    \multicolumn{2}{c}{\textbf{PubMed-OCR}} &
    \multicolumn{2}{c}{\textbf{AVG}} \\
    &  & CER & WER & CER & WER & CER & WER & CER & WER \\
    \midrule
    Qwen2.5-VL-3B  &    & 0.717 & 0.825 & 0.596 & 0.708 & 0.784 & 0.915 & 0.699 & 0.816 \\
    \quad GutenOCR-3B & 1  & 0.115 & 0.204 & 0.104 & 0.204 & 0.244 & 0.518 & 0.154 & 0.309 \\
    \quad GutenOCR-3B & 2  & 0.108 & 0.197 & 0.092 & 0.191 & 0.224 & 0.505 & 0.142 & 0.298 \\
    \quad GutenOCR-3B & 3a &
        \underline{\textbf{0.104}} & \underline{\textbf{0.192}} &
        \underline{\textbf{0.088}} & \underline{\textbf{0.190}} &
        0.134 & 0.220 &
        \underline{\textbf{0.109}} & \underline{\textbf{0.200}} \\
    \quad GutenOCR-3B & 3b & 0.129 & 0.362 & 0.112 & 0.344 & \underline{\textbf{0.113}} & \underline{\textbf{0.174}} & 0.118 & 0.283 \\
    \cmidrule{3-10}
    Qwen2.5-VL-7B  &    & 0.521 & 0.641 & 0.424 & 0.548 & 0.646 & 0.811 & 0.530 & 0.667 \\
    \quad GutenOCR-7B & 1  & 0.122 & 0.208 & 0.128 & 0.231 & 0.262 & 0.532 & 0.171 & 0.324 \\
    \quad GutenOCR-7B & 2  &
        \underline{0.109} & \underline{0.196} &
        \underline{0.097} & \underline{0.196} &
        0.241 & 0.513 &
        0.149 & 0.302 \\
    \quad GutenOCR-7B & 3a & 0.110 & 0.197 & 0.100 & 0.203 & 0.178 & 0.269 & \underline{0.129} & \underline{0.223} \\
    \quad GutenOCR-7B & 3b & 0.147 & 0.406 & 0.123 & 0.369 & \underline{0.137} & \underline{0.197} & 0.136 & 0.324 \\
    \bottomrule
    \end{tabular}
    \caption{
        Localized reading from ground-truth regions.
        We report CER and WER, lower is better ($\downarrow$).
        Bold entries are best overall per column; underlined entries are best within each backbone cohort.
    }
    \label{tab:localized-reading}
\end{table}

\paragraph{Localized reading from ground-truth regions.}

Table~\ref{tab:localized-reading} evaluates recognition given ground-truth boxes.
GutenOCR dramatically improves both backbones: average CER drops from $\approx$0.7/0.53 to the 0.11--0.20 range and WER from $\approx$0.8/0.67 to 0.20--0.32.
Unlike full-page reading, GutenOCR-3B Stage~3a achieves the best overall CER and WER, with GutenOCR-7B stages close behind.
Stage~3b again tilts toward PubMed-OCR, improving that domain while slightly degrading business-document performance.

\subsubsection{Detection}
\label{app:detailed-evaluation-detection}

\begin{table}[htb]
    \centering\small
    \begin{tabular}{lc cc cc cc cc}
    \toprule
    \multirow{2}{*}{\textbf{Model}} &
    \multirow{2}{*}{\textbf{Stage}} &
    \multicolumn{2}{c}{\textbf{OCR-IDL}} &
    \multicolumn{2}{c}{\textbf{TabMe++}} &
    \multicolumn{2}{c}{\textbf{PubMed-OCR}} &
    \multicolumn{2}{c}{\textbf{AVG}} \\
     & & F1 & R & F1 & R & F1 & R & F1 & R \\
    \midrule
    Qwen2.5-VL-3B  &    & 0.103 & 0.097 & 0.253 & 0.235 & 0.048 & 0.044 & 0.135 & 0.125 \\
    \quad GutenOCR-3B & 1  &
        \underline{\textbf{0.820}} & \underline{\textbf{0.812}} &
        0.795 & 0.761 &
        0.782 & 0.771 &
        \underline{\textbf{0.799}} & 0.794 \\
    \quad GutenOCR-3B & 2  & 0.800 & 0.798 & 0.806 & 0.815 & 0.758 & 0.754 & 0.788 & 0.789 \\
    \quad GutenOCR-3B & 3a &
        0.794 & 0.794 &
        \underline{\textbf{0.813}} & \underline{\textbf{0.821}} &
        0.788 & 0.787 &
        0.798 & \underline{\textbf{0.801}} \\
    \quad GutenOCR-3B & 3b &
        0.756 & 0.769 &
        0.798 & 0.809 &
        \underline{\textbf{0.817}} & \underline{\textbf{0.817}} &
        0.790 & 0.798 \\
    \cmidrule{3-10}
    Qwen2.5-VL-7B  &    & 0.117 & 0.114 & 0.162 & 0.161 & 0.054 & 0.056 & 0.111 & 0.110 \\
    \quad GutenOCR-7B & 1  &
        \underline{0.817} & \underline{0.810} &
        0.778 & 0.777 &
        0.775 & 0.769 &
        \underline{0.790} & 0.785 \\
    \quad GutenOCR-7B & 2  & 0.806 & 0.809 & 0.793 & 0.804 & 0.735 & 0.742 & 0.778 & 0.785 \\
    \quad GutenOCR-7B & 3a &
        0.794 & 0.800 &
        \underline{0.801} & \underline{0.814} &
        0.767 & 0.770 &
        0.787 & \underline{0.795} \\
    \quad GutenOCR-7B & 3b &
        0.720 & 0.735 &
        0.764 & 0.782 &
        \underline{0.799} & \underline{0.809} &
        0.761 & 0.775 \\
    \bottomrule
    \end{tabular}
    \caption{
        Full-page line detection.
        We report F1 and recall (R), higher is better ($\uparrow$).
        Bold numbers indicate the best value across all models; underlined numbers are best within each backbone cohort.
    }
    \label{tab:detection-lines}
\end{table}

\paragraph{Full-page line detection.}

Table~\ref{tab:detection-lines} shows that the raw Qwen2.5-VL models are very weak detectors (F1 $\approx$ 0.1).
GutenOCR raises both F1 and recall to around 0.75--0.82 across datasets for both 3B and 7B, turning the backbones into strong line detectors with only small size-dependent differences.
Stage~1 offers the best general-purpose detector, Stage~3a slightly favors TabMe++, and Stage~3b leans toward PubMed-OCR, but all GutenOCR stages are a large step up from the prompting-only baselines.

\subsubsection{Conditional Detection}
\label{app:detailed-evaluation-conditional}

\begin{table}[htb]
    \centering\small
    \begin{tabular}{lc cc cc cc cc}
    \toprule
    \multirow{2}{*}{\textbf{Model}} &
    \multirow{2}{*}{\textbf{Stage}} &
    \multicolumn{2}{c}{\textbf{OCR-IDL}} &
    \multicolumn{2}{c}{\textbf{TabMe++}} &
    \multicolumn{2}{c}{\textbf{PubMed-OCR}} &
    \multicolumn{2}{c}{\textbf{AVG}} \\
     & & F1 & R & F1 & R & F1 & R & F1 & R \\
    \midrule
    Qwen2.5-VL-3B  &    & 0.089 & 0.107 & 0.255 & 0.300 & 0.019 & 0.024 & 0.121 & 0.144 \\
    \quad GutenOCR-3B & 1  &
        \underline{0.861} & \underline{0.863} &
        0.882 & 0.884 &
        0.819 & 0.820 &
        0.854 & 0.855 \\
    \quad GutenOCR-3B & 2  & 0.855 & 0.858 & 0.898 & 0.901 & 0.799 & 0.799 & 0.851 & 0.853 \\
    \quad GutenOCR-3B & 3a &
        0.859 & 0.861 &
        \underline{\textbf{0.903}} & \underline{\textbf{0.905}} &
        0.868 & 0.870 &
        \underline{0.877} & \underline{0.879} \\
    \quad GutenOCR-3B & 3b &
        0.831 & 0.834 &
        0.888 & 0.894 &
        \underline{0.884} & \underline{0.887} &
        0.868 & 0.872 \\
    \cmidrule{3-10}
    Qwen2.5-VL-7B  &    & 0.307 & 0.326 & 0.466 & 0.489 & 0.083 & 0.090 & 0.285 & 0.302 \\
    \quad GutenOCR-7B & 1  & 0.874 & 0.875 & 0.887 & 0.889 & 0.852 & 0.852 & 0.871 & 0.872 \\
    \quad GutenOCR-7B & 2  &
        \underline{\textbf{0.880}} & \underline{\textbf{0.881}} &
        0.899 & 0.899 &
        0.849 & 0.849 &
        0.876 & 0.876 \\
    \quad GutenOCR-7B & 3a &
        0.873 & 0.874 &
        \underline{0.902} & \underline{0.904} &
        0.872 & 0.872 &
        \underline{\textbf{0.882}} & \underline{\textbf{0.883}} \\
    \quad GutenOCR-7B & 3b &
        0.813 & 0.815 &
        0.876 & 0.880 &
        \underline{\textbf{0.908}} & \underline{\textbf{0.911}} &
        0.866 & 0.869 \\
    \bottomrule
    \end{tabular}
    \caption{
        Conditional line detection given a query string.
        We report F1 and recall (R), higher is better ($\uparrow$).
        Bold entries are best overall per column; underlined entries are best within backbone.
    }
    \label{tab:conditional-detection}
\end{table}

\paragraph{Phrase-conditioned detection.}

Table~\ref{tab:conditional-detection} summarizes phrase-conditioned detection (“where is string $q$ on this page?”).
While Qwen2.5-VL-7B is noticeably better than 3B out of the box, both backbones are far from usable.
GutenOCR pushes F1 and recall into the 0.85--0.90 range across datasets, with Stage~3a giving the best average performance and Stage~3b again specializing toward PubMed-OCR.
In this setting the 7B model keeps a small but consistent edge over 3B, reflecting its stronger underlying recognition once detection has been trained.

\end{document}